\documentclass[runningheads]{llncs}
\usepackage{graphicx}
\usepackage{amsmath,amssymb} % define this before the line numbering.
\usepackage{color}
\usepackage[width=122mm,left=12mm,paperwidth=146mm,height=193mm,top=12mm,paperheight=217mm]{geometry}

\usepackage[breaklinks=true,letterpaper=true,colorlinks,bookmarks=false]{hyperref}
\usepackage{float}
\usepackage{booktabs}
\usepackage{multirow}

\usepackage{cite}

\renewcommand{\v}{\text{vec}}
\renewcommand{\a}{\mathcal{A}}

\newcommand{\etal}{\emph{et al. }}
\newcommand{\eg}{\emph{e.g. }}
\newcommand{\ie}{\emph{i.e. }}
\newcommand{\wrt}{\emph{w.r.t. }}

\begin{document}
% \linenumbers
\pagestyle{headings}
\mainmatter

\title{Symmetric Non-Rigid Structure from Motion for Category-Specific Object Structure Estimation} % Replace with your title

\titlerunning{Symmetric NRSfM for Category-Specific Object Structure Estimation}

\authorrunning{Y. Gao and A. Yuille}

\author{Yuan Gao$^1$\thanks{This work was been done when Yuan Gao was a visiting student in UCLA.} and Alan L. Yuille$^{2,3}$}
\institute{$^1$ City University of Hong Kong \quad $^2$ UCLA \quad $^3$ John Hopkins University \\
\email{ \{Ethan.Y.Gao, Alan.L.Yuille\}@gmail.com}
}

\maketitle

\begin{abstract}
Many objects, especially these made by humans, are symmetric, \eg cars and aeroplanes. This paper addresses the estimation of 3D structures of symmetric objects from multiple images of the same object category, \eg different cars, seen from various viewpoints. We assume that the deformation between different instances from the same object category is non-rigid and symmetric. In this paper, we extend two leading non-rigid structure from motion (SfM) algorithms to exploit symmetry constraints. We model the both methods as energy minimization, in which we also recover the missing observations caused by occlusions. In particularly, we show that by rotating the coordinate system, the energy can be decoupled into two independent terms, which still exploit symmetry, to apply matrix factorization separately on each of them for initialization. The results on the Pascal3D+  dataset show that our  methods significantly improve performance over baseline methods.

\end{abstract}

% \vspace{-9mm}
\section{Introduction}
% \vspace{-1mm}

3D structure reconstruction is a major task in computer vision. Structure from motion (SfM) method, which aims at  estimating the 3D structure by the 2D annotated keypoints from 2D image sequences, has been proposed for rigid objects \cite{Tomasi92}, and was later extended to non-rigidity \cite{Hartley2004,Torresani03,Xiao04,Torresani08,Akhter2011,Gotardo11,Hamsici2012,Dai12,Dai14,ma2015non,ma2013robust,ma2013regularized,Agudo14}. Many man-made objects have symmetric structures \cite{Rosen12,Hong04}. Motivated by this, symmetry has been studied extensively in the past decades \cite{Gordon90,kontsevich93,Vetter94,Mukherjee95,Hong04,Thrun05,Li07}. However, this information has not been exploited in recent works on 3D object reconstruction \cite{Vicente14,Kar15}, nor used in standard non-rigid structure from motion (NRSfM) algorithms \cite{Torresani03,Xiao04,Torresani08,Akhter2011,Gotardo11,Hamsici2012,Dai12,Dai14,Agudo14}.

The goal of this paper is to investigate how symmetry can improve NRSfM. Inspired by recent works \cite{Vicente14,Kar15}, we are interested in estimating the 3D structure of objects, such as cars, airplanes, \emph{etc}. This differs from the classic SfM problem because our input are images of several different object instances from the same category (\eg different cars), instead of sequential images of the same object undergoing motion. In other words, our goal is to estimate the 3D structures of objects from the same class, given intra-class instances from various viewpoints. Specifically, the Pascal3D+ keypoint annotations on different objects from the same category are used as input to our method, where the the symmetric keypoint pairs can also be easily inferred. In this paper, non-rigidity means the deformation between the objects from same category can be non-rigid, \eg between sedan and SUV cars, but the objects themselves are rigid and symmetric.

By exploiting symmetry, we propose two symmetric NRSfM methods. By assuming that the 3D structure can be represented by a linear combination of basis functions (the coefficients vary for different objects): one method is an extension of \cite{Torresani08} which is based on an EM approach with a Gaussian prior on the coefficients of the deformation bases, named Sym-EM-PPCA; the other method, \ie Sym-PriorFree, is an extension of \cite{Dai12,Dai14}, which is a direct matrix factorization method without prior knowledge. For fair comparison, we use the same projection models and other assumptions used in \cite{Torresani08} and \cite{Dai12,Dai14}.

More specifically, our Sym-EM-PPCA method, following \cite{Torresani08}, assumes weak  perspective projection (\ie the orthographic projection plus scale). We group the keypoints into symmetric keypoint pairs. We assume that the 3D structure is also symmetric and consists of a mean shape (of that category) and a linear combination of the symmetric deformation bases. As in \cite{Torresani08}, we put a Gaussian prior on the coefficient of the deformation bases. This is intended partly to regularize the problem and partly to deal an apparent ambiguity in non-rigid structure from motion. But recent work \cite{Akhter09} showed that this is a ``gauge freedom'' which does not affect the estimation of 3D structure, so the prior is not really needed.

Our Sym-PriorFree method is based on prior free non-rigid SfM algorithms \cite{Dai12,Dai14}, which build on the insights in \cite{Akhter09}. We formulate the problem of estimating 3D structure and camera parameters in terms of minimizing an energy function, which exploits symmetry, and at the same time can be re-expressed as the sum of two independent energy functions. Each of these energy functions can be minimized separately by matrix factorization, similar to the methods in \cite{Dai12,Dai14}, and the ambiguities are resolved using orthonormality constraints on the viewpoint parameters. This extends work in a companion paper \cite{Gao2016_Rigid}, which shows how symmetry can be used to improve rigid structure from motion methods \cite{Tomasi92}.

Our main contributions are: (I) Sym-EM-PPCA, which imposes symmetric constraints on both 3D structure and deformation bases. Sym-Rigid-SfM (see our companion paper \cite{Gao2016_Rigid}) is used to initialize Sym-EM-PPCA with hard symmetric constraints on the 3D structure. (II) Sym-PriorFree, which extends the matrix factorization methods of \cite{Dai12,Dai14}, to initialize a coordinate descent algorithm.

In this paper, we group keypoints into symmetric keypoint pairs, and use a superscript $\dag$ to denote symmetry, \ie $Y$ and $Y^{\dag}$ are the 2D symmetric keypoint pairs. The  paper is organized as follows: firstly, we review related works in Section 2. In Section 3, the ambiguities in non-rigid SfM are discussed. Then we present the Sym-EM-PPCA algorithm and Sym-PriorFree algorithm in Section 4. After that, following the experimental settings in \cite{Kar15}, we evaluated our methods on the  Pascal3D+ dataset \cite{Xiang14} in Section 5. Section 5 also includes diagnostic results on the noisy 2D annotations to show that our methods are robust to imperfect symmetric annotations.  Finally, we give our conclusions in Section 6.

% \vspace{-9mm}
\section{Related Works}

There is a long history of using symmetry as a cue for computer vision tasks. For example, symmetry has been used in depth recovery \cite{Gordon90,kontsevich93,Mukherjee95} as well as recognizing symmetric objects \cite{Vetter94}. Several geometric clues, including symmetry, planarity, orthogonality and parallelism have been taken into account for 3D scene reconstruction \cite{grossmann2002maximum,grossmann2005least}, in which the author used pre-computed camera rotation matrix by vanishing point \cite{grossmann2002single}. Recently, symmetry has been applied in more areas such as 3D mesh reconstruction with occlusion \cite{Thrun05}, and scene reconstruction \cite{Hong04}. For  3D keypoints reconstruction, symmetry, incorporated with planarity and compactness prior, has also been studied in \cite{Li07}.

SfM has also been studied extensively in the past decades, ever since the seminal work on rigid SfM \cite{kontsevich87,Tomasi92}. Bregler \etal extended this  to the non-rigid case \cite{Bregler00}. A Column Space Fitting (CSF) method was proposed for rank-$r$ matrix factorization (MF) for SfM with smooth time-trajectories assumption \cite{Gotardo11}, which was later unified in a more general MF framework \cite{hongsecrets}\footnote{However, the general framework in \cite{hongsecrets} cannot be used to SfM directly, because they did not constrain that all the keypoints have the same translation.
%The main focuses in \cite{hongsecrets} are better optimization of rank-$r$ matrix factorization and better runtime.
}. Early analysis of NRSfM showed that there were ambiguities in 3D structure reconstruction \cite{Xiao04}. This lead to studies which assumed priors on the NR deformations \cite{Xiao04,Torresani08,Olsen08,Akhter08,Gotardo11,Akhter2011}. But it was then shown that these ambiguities did not affect the final estimate of 3D structure, \ie all legitimate solutions lying in the same subspace (despite under-constrained) give the same solutions for the 3D structure \cite{Akhter09}. This facilitated the invention of prior free matrix factorization method for NRSfM \cite{Dai12,Dai14}. Recently SfM methods have also been used for category-specific object reconstruction, \eg estimating the shape of different \emph{cars} under various viewing conditions \cite{Kar15,Vicente14}, but the symmetry cues was not exploited. Note that repetition patterns have recently been incorporated into SfM for urban facades reconstruction \cite{Ceylan14}, but this mainly focused on repetition detection and registration. Finally, in a companion paper \cite{Gao2016_Rigid}, we exploited symmetry for rigid SfM.

\section{The Ambiguities in Non-rigid SfM}

This section reviews the intrinsic ambiguities in non-rigid SfM, \ie (i) the ambiguities between the camera projection and the 3D structure, and (ii) the ambiguities between  the deformation bases and their coefficients \cite{Akhter09}.
In the following sections (\ie in Remark \ref{Remark3}), we will show the ambiguity between camera projection and 3D structure (\ie originally the $3 \times 3$ matrix ambiguity as discussed below) can be decomposed into two types of ambiguities under the symmetric constraints, \ie a scale ambiguity along the symmetry axis, and a $2 \times 2$ matrix ambiguity on the other two axes.

The key idea of non-rigid SfM is to represent the non-rigid deformations of objects in terms of a linear combination of bases:
\begin{equation}  \mathbf{Y} = \mathbf{R} \mathbf{S}  \quad \text{and} \quad  \mathbf{S} = \mathbf{V} \mathbf{z}, \ \ \mathbf{R} \mathbf{R}^T = I,\end{equation}
where $\mathbf{Y}$ is the stacked 2D keypoints, $\mathbf{R}$ is the camera projection for the $N$ images. $\mathbf{S}$ is the 3D structure which is modeled by the linear combination of the stacked deformation bases $\mathbf{V}$, and $\mathbf{z}$ is the coefficient.

Firstly, as is well known, there are ambiguities between the projection $\mathbf{R}$ and the 3D structure $\mathbf{S}$ in the matrix factorization, \ie let $\mathbf{A}_1$ be an invertible matrix, then $\mathbf{R} \leftarrow \mathbf{R} \mathbf{A}_1$ and $\mathbf{S} \leftarrow \mathbf{A}^{-1}_1 \mathbf{S}$ will not change the value of $\mathbf{R} \mathbf{S}$. These ambiguities can be solved by imposing orthogonality constraints on the camera parameters $\mathbf{R} \mathbf{R}^T = I$ up to a fixed rotation, which is a ``gauge freedom'' \cite{morris2001gauge} corresponding to a choice of coordinate system.

In addition, there are other ambiguities between the coefficients $\mathbf{z}$ and the deformation bases $\mathbf{V}$ \cite{Xiao04}. Specifically, let $\mathbf{A}_2$ be another invertible matrix, and let $\mathbf{w}$ lie in the null space of the projected deformation bases $\mathbf{R} \mathbf{V}$, then $\mathbf{z} \leftarrow \mathbf{A}_2 \mathbf{z} $ and $\mathbf{V} \leftarrow  \mathbf{V}\mathbf{A}^{-1}_2$, or $\mathbf{z} \leftarrow \mathbf{z} + \alpha \mathbf{w}$ will not change the value of $\mathbf{R} \mathbf{V} \mathbf{z}$. This motivated Bregeler {\it et al.} to impose a Gaussian prior on the coefficient $\mathbf{z}$ in order to eliminate the ambiguities. Recently, it was proved in \cite{Akhter09} that these ambiguities are also ``fake'', \ie they do not affect the estimate of the 3D structure. This proof facilitated prior-free matrix factorization methods for  non-rigid SfM \cite{Dai12,Dai14}.

\section{Symmetric Non-Rigid Structure from Motion}

In this paper we extend non-rigid SfM methods by requiring that the 3D structure is symmetric. We assume the deformations are non-rigid and also symmetric\footnote{We assume symmetric deformations because our problem involves deformations from one symmetric object to another. But it also can be extended to non-symmetric deformations straightforwardly.}.
We propose two symmetric non-rigid SfM models. One is the extension of the iterative EM-PPCA model with a prior on the deformation coefficients \cite{Torresani08}, and the other extends the prior-free matrix factorization model \cite{Dai12,Dai14}.

For simplicity of derivation, we focus on estimating the 3D structure and camera parameters. In practice, there are occluded keypoints in almost all images in the Pascal3D+ dataset. But we use standard ways to deal with them, such as initializing them ignoring symmetry by rank 3 recovery using the first 3 largest singular value, then treating them as missing data to be estimated by EM or coordinate descent algorithms. In our companion paper \cite{Gao2016_Rigid}, we gave details of these methods for the simpler case of rigid structure from motion.

Note that we use slightly different camera models for Sym-EM-PPCA (weak perspective projection) and Sym-PriorFree (orthographic projection). This is to stay consistent with the non-symmetric methods which we compare with, namely \cite{Torresani08} and \cite{Dai12,Dai14}. Similarly, we treat translation differently by either centeralizing the data or treating it as a variable to be estimated, as appropriate. We will discuss this further when  presenting the Sym-PriorFree method.

\subsection{The Symmetric EM-PPCA Model}

In EM-PPCA \cite{Torresani08}, Bregler \etal assume that the 3D structure is represented by a mean structure $\bar{S}$ plus a non-rigid deformation. Suppose there are $P$ keypoints on the structure, the non-rigid model of EM-PPCA is:
\begin{equation}
\mathbb{Y}_n = G_n (\mathbb{\bar{S}} + \mathbf{V}z_n) + \mathbb{T}_n + N_n, \label{EM-PPCA}
\end{equation}
where $\mathbb{Y}_n \in \mathbb{R}^{2P \times 1}, \mathbb{\bar{S}} \in \mathbb{R}^{3P \times 1}$, and $\mathbb{T}_n \in \mathbb{R}^{2P \times 1}$ are the stacked vectors of 2D keypoints, 3D mean structure and translations. $G_n = I_{P} \otimes c_n R_n$, in which $c_n$ is the scale parameter for weak perspective projection, $\mathbf{V} = [\mathbb{V}_1, ..., \mathbb{V}_K] \in \mathbb{R}^{3P \times K}$ is the grouped $K$ deformation bases, $z_n \in \mathbb{R}^{K \times 1}$ is the coefficient of the $K$ bases, and $N_n$ is the Gaussian noise $N_n \sim \mathcal{N}(0, \sigma^2I)$.

Extending Eq. \eqref{EM-PPCA} to our symmetry problem in which there are $P$ keypoint pairs $\mathbb{Y}_n$ and $\mathbb{Y}_n^{\dag}$, we have:
\begin{equation}
\mathbb{Y}_n = G_n (\mathbb{\bar{S}} + \mathbf{V}z_n) + \mathbb{T}_n + N_n, \quad \mathbb{Y}_n^{\dag} = G_n (\mathbb{\bar{S}}^{\dag} + \mathbf{V}^{\dag}z_n) + \mathbb{T}_n + N_n. \label{3}
\end{equation}

Assuming that the object is symmetric along the $x$-axis, the relationship between $\mathbb{\bar{S}}$ and $\mathbb{\bar{S}}^{\dag}$, $\mathbf{V}$ and $\mathbf{V}^{\dag}$ are:
\begin{equation}
\mathbb{\bar{S}}^{\dag} = \a_P \mathbb{\bar{S}}, \qquad \mathbf{V}^{\dag} = \a_P \mathbf{V}, \label{Vrelation}
\end{equation}
where $\a_P = I_{P} \otimes \a$, $\a = \text{diag}([-1, 1, 1])$ is a matrix operator which negates the first row, and $I_P \in \mathbb{R}^{P \times P}$ is an identity matrix. Thus, we have\footnote{We set hard constraints on $\mathbb{\bar{S}}$ and $\mathbb{\bar{S}}^{\dag}$, \ie replace $\mathbb{\bar{S}}^{\dag}$ by $\a_P \mathbb{\bar{S}}$ in Eq. \eqref{EM-PPCA-Prob}, because it can be guaranteed by the Sym-RSfM initialization in our companion paper \cite{Gao2016_Rigid}. While the initialization on $\mathbf{V}$ and $\mathbf{V}^{\dag}$ by PCA cannot guarantee such a desirable property, thus a Language multiplier term is used for the constraint on $\mathbf{V}$ and $\mathbf{V}^{\dag}$ in Eq. \eqref{likelihood}.}:
\begin{align}
  &P(\mathbb{Y}_n | z_n, G_n, \mathbb{\bar{S}}, \mathbf{V}, \mathbb{T}) = \mathcal{N}(G_n (\mathbb{\bar{S}} + \mathbf{V}z_n) + \mathbb{T}_n, \sigma^2 I) \nonumber \\
  &P(\mathbb{Y}_n^{\dag} | z_n, G_n, \mathbb{\bar{S}}, \mathbf{V}^{\dag}, \mathbb{T}) = \mathcal{N}(G_n (\a_P \mathbb{\bar{S}} + \mathbf{V}^{\dag}z_n) + \mathbb{T}_n, \sigma^2 I) \label{EM-PPCA-Prob}
\end{align}

Following Bregler \etal \cite{Torresani08}, we introduce a prior $P(z_n)$ on the coefficient variable $z_n$. This prior is a zero mean unit variance Gaussian. It is used for (partly) regularizing the inference task but also for dealing with the ambiguities between basis coefficients $z_n$ and bases $\mathbf{V}$, as mentioned above (when \cite{Torresani08} was published it was not realized that these are ``gauge freedom''). This enables us to treat $z_n$ as the hidden variable and use EM algorithm to estimate the structure and camera viewpoint parameters. The formulation of the problem, in terms of Gaussian distributions (or, more technically, the use of conjugate priors) means that both steps of the EM algorithm are straightforward to implement.

\begin{remark}
Our Sym-EM-PPCA method is a natural extension of the method in \cite{Torresani08} to maximize the marginal probability $P(\mathbb{Y}_n, \mathbb{Y}_n^{\dag} | G_n, \mathbb{\bar{S}}, \mathbf{V}, \mathbf{V}^{\dag}, \mathbb{T})$ with a Gaussian prior on $z_n$ and a Language multiplier term (\ie a regularization term) on $\mathbf{V}, \mathbf{V}^{\dag}$. This can be solved by \emph{general EM} algorithm \cite{Bishop06}, where both the \textbf{E} and \textbf{M} steps take simple forms because the underlying probability distributions are Gaussians (due to conjugate Gaussian prior).
\end{remark}

\begin{figure}[!htp]
\setlength{\belowcaptionskip}{-10pt}
\centering
\includegraphics[width=0.3\linewidth]{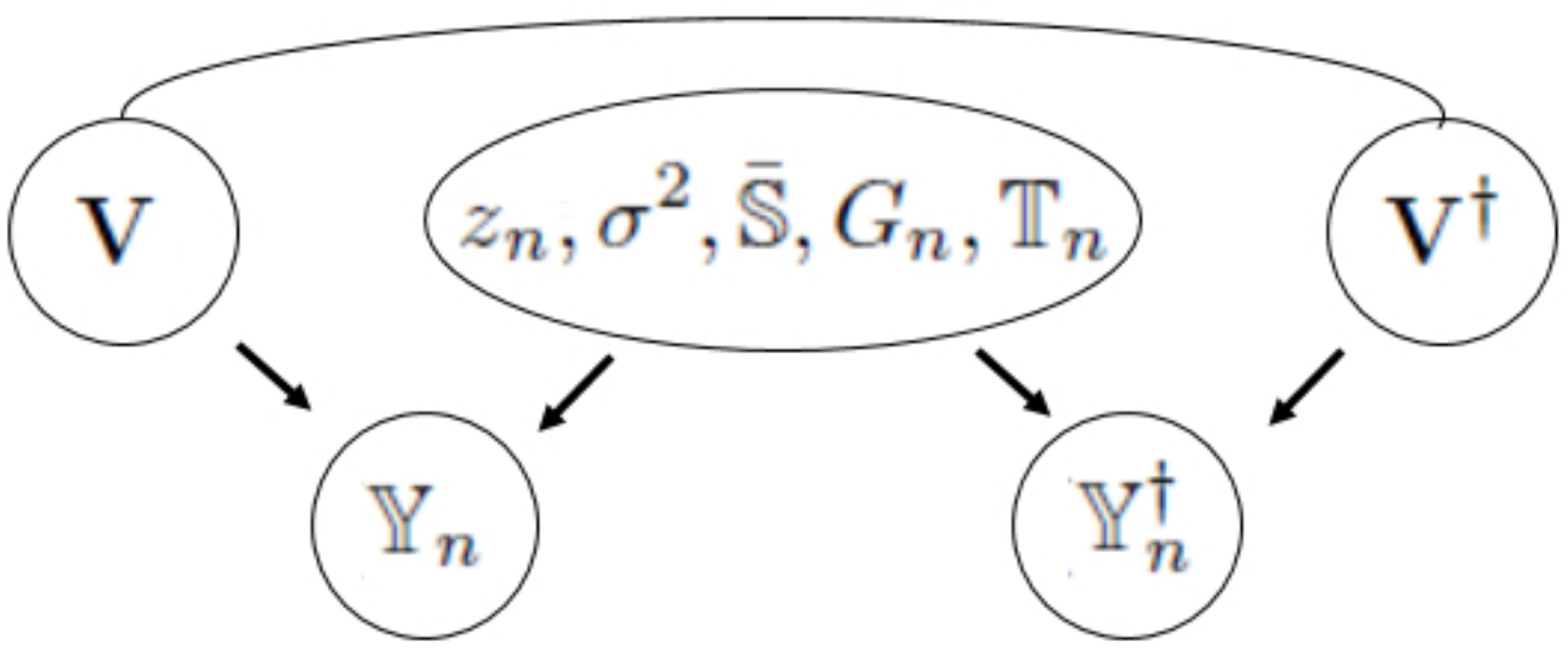}
\caption{The graphical model of the variables and parameters.}
\label{graphical}
\end{figure}

\textbf{E-Step}: This step is to get the statistics of $z_n$ from its posterior. Let the prior on $z_n$ be $P(z_n) = \mathcal{N}(0,I)$ as in \cite{Torresani08}. Then, we have $P(z_n)$, $P(\mathbb{Y}_n | z_n; \sigma^2, \mathbb{\bar{S}}, \mathbf{V}, G_n, \mathbb{T}_n)$ and $P(\mathbb{Y}_n^{\dag} | z_n; \sigma^2, \mathbb{\bar{S}}, \mathbf{V}^{\dag}, G_n, \mathbb{T}_n)$, which do not provide the complete posterior distribution directly. Fortunately, the conditional dependence of the variables shown in Fig. \ref{graphical} (graphical model) implies that the posterior of $z_n$ can be calculated by:
\begin{align}
& P(z_n|\mathbb{Y}_n, \mathbb{Y}_n^{\dag}; \sigma^2, \mathbb{\bar{S}}, \mathbf{V}, \mathbf{V}^{\dag}, G_n, \mathbb{T}_n) \nonumber \\
\sim& P(z_n, \mathbb{Y}_n, \mathbb{Y}_n^{\dag}| \sigma^2, \mathbb{\bar{S}}, \mathbf{V}, \mathbf{V}^{\dag}, G_n, \mathbb{T}_n) \nonumber \\
=& P(\mathbb{Y}_n | z_n; \sigma^2, \mathbb{\bar{S}}, \mathbf{V}, G_n, \mathbb{T}_n) P(\mathbb{Y}_n^{\dag} | z_n; \sigma^2, \mathbb{\bar{S}}, \mathbf{V}^{\dag}, G_n, \mathbb{T}_n) P(z_n) \nonumber \\
=& \mathcal{N}(z_n | \mu_n, \Sigma_n) \label{z_n_bregler}
\end{align}
The last equation of Eq. \eqref{z_n_bregler} is obtained by the fact that the prior and the conditional distributions of $z_n$ are all Gaussians (conjugate prior). Then the first and second order statistics of $z_n$ can be obtained as:
\begin{align}
&\mu_n = \gamma \left\{ \mathbf{V}^TG_n^T(\mathbb{Y}_n-G_n\mathbb{\bar{S}}-\mathbb{T}_n) + \mathbf{V}^{\dag T}G_n^T(\mathbb{Y}_n^{\dag}-G_n\a_P\mathbb{\bar{S}}-\mathbb{T}_n) \right\} \label{8}\\
&\phi_n = \sigma^2\gamma ^{-1} + \mu_n\mu_n^T \label{9}
\end{align}
where $\gamma = (\mathbf{V}^TG_n^TG_n\mathbf{V} + \mathbf{V}^{\dag T}G_n^TG_n\mathbf{V}^{\dag} + \sigma^2I)^{-1}$.

\textbf{M-Step}: This is to maximize the joint likelihood which is similar to the coordinate descent in Sym-RSfM (in a companion paper \cite{Gao2016_Rigid}) and that in Sym-PriorFree method in the later sections. The complete log-likelihood $Q(\theta)$ is:
{\small
\begin{align}
\mathcal{Q}(\theta) =& -\sum_n\ln P(\mathbb{Y}_n, \mathbb{Y}_n^{\dag} | z_n; G_n, \mathbb{\bar{S}}, \mathbf{V},  \mathbf{V}^{\dag}, \mathbb{T}) + \lambda||\mathbf{V}^{\dag} - \a_P\mathbf{V} ||^2 \nonumber \\
=& -\sum_n \left( \ln P(\mathbb{Y}_n | z_n; G_n, \mathbb{\bar{S}}, \mathbf{V}, \mathbb{T}) + \ln P(\mathbb{Y}_n^{\dag} | z_n; G_n, \mathbb{\bar{S}}, \mathbf{V}^{\dag}, \mathbb{T})\right) + \lambda||\mathbf{V}^{\dag} - \a_P\mathbf{V} ||^2 \nonumber \\
&\text{s. t.} \qquad R_n R_n^T = I, \qquad \text{ where } \theta = \{G_n, \mathbb{\bar{S}}, \mathbf{V}, \mathbf{V}^{\dag}, \mathbb{T}_n, \sigma^2 \}. \label{likelihood}
\end{align}
}
~~ The maximization of Eq. \eqref{likelihood} is straightforward, \ie taking the derivative of each unknown parameter in $\theta$ and equating it to 0. The update rule of each parameter is very similar to the original EM-PPCA \cite{Torresani08} (except $\mathbb{\bar{S}}, \mathbf{V}, \mathbf{V}^{\dag}$ should be updated jointly), which we put in Appendix \ref{sect:M}.

\emph{Initialization.} $\mathbf{V}$ and $\mathbf{V}^{\dag}$ are initialized by the PCA on the residual of the 2D keypoints minus their rigid projections iteratively. Other variables (including the rigid projections) are initialized by Sym-RSfM \cite{Gao2016_Rigid}. Specifically, $R_n$, $\bar{S}$ and the occluded points $Y_{n,p}, Y_{n,p}^{\dag}$ can be initialized directly from Sym-RSfM, $c_n$ is initialized as 1, $t_n$ is initialized by $t_n = \sum_p (Y_{n,p} - R_n \bar{S}_p + Y_{n,p}^{\dag} - R_n \mathcal{A} \bar{S}_p)$.

\subsection{The Symmetric Prior-Free Matrix Factorization Model} \label{SPF}

In the Prior-Free NRSfM \cite{Dai12,Dai14}, Dai \etal also used the linear combination of several deformations bases to represent the non-rigid deformation. But, unlike EM-PPCA \cite{Torresani08}, Dai \etal estimated the non-rigid structure directly without using the mean structure and the prior on the coefficients. We make the same assumptions so that we can directly compare with them.

Assume that $Y_n \in \mathbb{R}^{2 \times P}$ are the $P$ keypoints for image $n$, then we have:
\begin{align}
& Y_n = R_n S_n = [z_{n1} R_n, ..., z_{nK} R_n] [\mathbf{V}_1, ..., \mathbf{V}_K]^T = \Pi_n \mathbf{V}, \nonumber \\
& Y_n^{\dag} = R_n S_n^{\dag} = [z_{n1} R_n, ..., z_{nK} R_n] [\mathbf{V}_1^{\dag}, ..., \mathbf{V}_K^{\dag}]^T = \Pi_n \mathbf{V}^{\dag},
\label{PF0}
\end{align}
where $\mathbf{z}_n = [z_{n1}, ..., z_{nK}] \in \mathbb{R}^{1 \times K}$, $\Pi_n = R_n (\mathbf{z}_n \otimes I_3) \in \mathbb{R}^{2 \times 3K}$, and $\mathbf{V} = [\mathbf{V}_1^T, ..., \mathbf{V}_K^T]^T \in \mathbb{R}^{3K \times P}$.

Without loss of generality, we assume that the symmetry is across the $x$-axis: $S_n = \mathcal{A} S_n^{\dag}$, where $\mathcal{A} = \text{diag}[-1, 1, 1]^T$ is a matrix operator negating the first row of $S_n$. Then the first two terms in Eq. \eqref{PF0} give us the energy function (or the likelihood) to estimate the unknown $R_n, S_n$ and recover the missing data by \emph{coordinate descent} on:
\begin{align}
&\mathcal{Q}(R_n, S_n, \{Y_{n,p}, (n,p) \in {\it IVS}\}, \{Y_{n,p}^{\dag}, (n,p) \in {\it IVS}^{\dag}\}) \nonumber \\
=&  \sum_{(n,p) \in {\it VS}} ||Y_{n,p} - R_n S_{n, p}||^2_2 + \sum_{(n,p) \in {\it VS}^{\dag}}||Y_{n,p}^{\dag} - R_n \a S_{n, p} ||^2_2 + \nonumber \\
& \sum_{(n,p) \in I{\it VS}} ||Y_{n,p} - R_n S_{n, p}||^2_2 + \sum_{(n,p) \in {\it IVS}^{\dag}}||Y_{n,p}^{\dag} - R_n \a S_{n, p} ||^2_2, \label{PF1}
\end{align}
where $VS$ and $IVS$ are the index sets of the \emph{visible} and \emph{invisible} keypoints, respectively. $Y_{n,p}$ and $S_{n,p}$ are the 2D and 3D $p$'th keypoints of the $n$'th image. We treat the $\{Y_{n,p}, (n,p) \in I{\it VS}\},
\{Y_{n,p}^{\dag}, (n,p) \in I{\it VS}^{\dag}\}$ as missing/hidden variables to be estimated.

\begin{remark}
It is straightforward to minimize Eq. \eqref{PF1} by \emph{coordinate descent}. The missing points can be initialized simply by rank 3 recovery (\ie by the reconstruction using the first 3 largest singular value) ignoring the symmetry property and non-rigidity. But it is much harder to get good initializations for the $R_n$ and $S_n$. In the following, we will describe how we get good initializations for each $R_n$ and $S_n$ exploiting symmetry  after the missing points have been initialized.
\end{remark}

Let $\mathbf{Y}$ is the stacked keypoints of $N$ images, $\mathbf{Y} = [Y_1^T, ..., Y_N^T]^T \in \mathbb{R}^{2N \times P}$, the model is represented by:
\begin{equation}
\mathbf{Y} = \mathbf{RS} =
\begin{bmatrix}
R_1 S_1 \\
\vdots \\
R_N S_N
\end{bmatrix}
=
\begin{bmatrix}
z_{11} R_1, &..., &z_{1K} R_1 \\
\vdots &\ddots &\vdots \\
z_{N1} R_N, &..., &z_{NK} R_N
\end{bmatrix}
\begin{bmatrix}
\mathbf{V}_1 \\
\vdots \\
\mathbf{V}_K
\end{bmatrix}
= \mathbf{\Pi} \mathbf{V}, \label{PFmodel}
\end{equation}
where $\mathbf{R} = \text{blkdiag}([R_1, ..., R_N]) \in \mathbb{R}^{2N \times 3N}$ are the stacked camera projection matrices, in which blkdiag denotes block diagonal. $\mathbf{S} = [S_1^T, ..., S_N^T]^T \in \mathbb{R}^{3N \times P}$ are the stacked 3D structures. $\mathbf{\Pi} = \mathbf{R} (\mathbf{z} \otimes I_3) \in \mathbb{R}^{2N \times 3K}$, where $\mathbf{z} \in \mathbb{R}^{N \times K}$ are the stacked coefficients. Similar equations apply to $\mathbf{Y}^{\dag}$.

Note that $\mathbf{R} \in \mathbb{R}^{2N \times 3N}$, $\mathbf{V} \in \mathbb{R}^{3K \times P}$ are stacked differently than how they were stacked for the Sym-EM-PPCA method (\ie $\mathbf{R} \in \mathbb{R}^{2N \times 3}$, $\mathbf{V} \in \mathbb{R}^{3P \times K}$).  It is because now we have $N$ different $S_n$'s  (\ie $\mathbf{S} \in \mathbb{R}^{3N \times P}$), while there is only one $\bar{S}$ in the Sym-EM-PPCA method.

In the following, we assume the deformation bases are symmetric, which  ensures that the non-rigid structures are symmetric (\eg the deformation from \emph{sedan} to \emph{truck} is non-rigid and symmetric since \emph{sedan} and \emph{truck} are both symmetric). This yields an energy function:
\begin{align}
\mathcal{Q}(\mathbf{R}, \mathbf{S}) =& ||\mathbf{Y} - \mathbf{R}  \mathbf{S}||_2^2 + ||\mathbf{Y}^{\dag} - \mathbf{R} \a_N\mathbf{S}^{\dag}||_2^2 \nonumber \\
=& ||\mathbf{Y} - \mathbf{\Pi} \mathbf{V} ||_2^2 + ||\mathbf{Y}^{\dag} - \mathbf{\Pi} \a_K \mathbf{V}^{\dag} ||_2^2, \label{energyPFNR}
\end{align}
where $\a_N = I_N \otimes \a, \a_K = I_K \otimes \a$, and $\a = \text{diag}([-1,1,1])$.

\begin{remark}
Note that we cannot use the first equation of Eq. \eqref{energyPFNR} to solve $\mathbf{R}, \mathbf{S}$ directly (even if not exploiting symmetry), because $\mathbf{Y}$ and $\mathbf{Y}^{\dag}$ are of rank $min \{2N, 3K, P \}$ but estimating $\mathbf{R}, \mathbf{S}$ directly by SVD on $\mathbf{Y}$ and/or $\mathbf{Y}^{\dag}$ requires rank $3N$ matrix factorization. Hence we focus on the last equation of Eq. \eqref{energyPFNR} to get the initialization of $\mathbf{\Pi}, \mathbf{V}$ firstly. Then, $\mathbf{R}, \mathbf{S}$ can be updated by coordinate descent on the first equation of Eq. \eqref{energyPFNR} under \emph{orthogonality constraints} on $\mathbf{R}$ and \emph{low-rank} constraint on $\mathbf{S}$.

Observe that the last equation of Eq. \eqref{energyPFNR} cannot be optimized directly by SVD either, because they consist of two  terms which are not independent. In other words, the matrix factorizations of $\mathbf{Y}$ and $\mathbf{Y}^{\dag}$ do not give consistent estimations of $\mathbf{\Pi}$ and $\mathbf{V}$. Instead, we now discuss how to estimate $\mathbf{\Pi}$ and $\mathbf{V}$ by rotating the coordinate axes (to decouple the depended energy terms), performing matrix factorization, and using subspace intersection (to eliminate the ambiguities), which is an extension of the original prior-free method \cite{Dai12,Dai14} and our companion Sym-RSfM \cite{Gao2016_Rigid}.
\end{remark}

We first rotate coordinate systems (of $\mathbf{Y},\mathbf{Y}^{\dag}$) to obtain decoupled equations:
\begin{equation}
\mathbf{L} = \frac{\mathbf{Y} - \mathbf{Y}^{\dag}}{2} = \hat{\mathbf{\Pi}}^1 \hat{\mathbf{V}}_x \qquad \mathbf{M} = \frac{\mathbf{Y} + \mathbf{Y}^{\dag}}{2} = \hat{\mathbf{\Pi}}^2 \hat{\mathbf{V}}_{yz},
\end{equation}
where the two righthand sides of the equation depend on different components of $\hat{\mathbf{\Pi}}, \hat{\mathbf{V}}$. More specifically, by discarding the all 0 rows of the bases, $\hat{\mathbf{\Pi}}^1 \in \mathbf{R}^{2N \times K}$, $\hat{\mathbf{\Pi}}^2 \in \mathbf{R}^{2N \times 2K}$, $\hat{\mathbf{V}}_x \in \mathbf{R}^{K \times P}$, $\hat{\mathbf{V}}_{yz} \in \mathbf{R}^{2K \times P}$.

This yield two independent energies to be minimized separately by SVD:
\begin{equation}
\mathcal{Q}(\mathbf{\mathbf{\Pi}}, \mathbf{V}) = ||\mathbf{L} - \hat{\mathbf{\Pi}}^1 \hat{\mathbf{V}}_x||_2^2 + ||\mathbf{M} - \hat{\mathbf{\Pi}}^2 \hat{\mathbf{V}}_{yz}||_2^2 \label{decoupledEnergy}
\end{equation}
\begin{remark}
We have formulated Sym-PriorFree as minimizing two energy terms in Eq. \eqref{decoupledEnergy}, which consists of independent variables. This implies that we can solve them by matrix factorization on each energy term separately, which gives solutions for $\mathbf{\Pi} = \mathbf{R} (\mathbf{z} \otimes I_3)$ and for the basis vectors $\mathbf{V}$ up to an ambiguity $H$. It will be discussed more explicitly in the following and we will show how to use orthogonality of the camera parameters to partially solve for $H$.
\end{remark}

Solving Eq. \eqref{decoupledEnergy} by matrix factorization gives us solutions up to a matrix ambiguity $H$. More precisely,
there are ambiguity matrices $H^1, H^2$ between the true solutions $\mathbf{\Pi}^1, \mathbf{V}_x, \mathbf{\Pi}^2, \mathbf{V}_{yz}$ and the initial estimation output by matrix factorization $\hat{\mathbf{\Pi}}^1, \hat{\mathbf{V}}_x, \hat{\mathbf{\Pi}}^2, \hat{\mathbf{V}}_{yz}$:
\begin{equation}
\mathbf{L} = \mathbf{\Pi}^1 \mathbf{V}_x = \hat{\mathbf{\Pi}}^1 H^1 (H^1)^{-1} \hat{\mathbf{V}}_x  \qquad \mathbf{M} = \mathbf{\Pi}^2 \mathbf{V}_{yz} = \hat{\mathbf{\Pi}}^2 H^2 (H^2)^{-1} \hat{\mathbf{V}}_{yz}  \label{V_PF}
\end{equation}
where $H^1 \in \mathbb{R}^{K \times K}$ and $H^2 \in \mathbb{R}^{2K \times 2K}$.

Now, the problem becomes to find $H^1, H^2$. Note that we have orthonormality constraints on each camera projection matrix $R_n$, which further impose constraints on $\Pi_n$. Thus, it can be used to partially estimate the ambiguity matrices $H^1, H^2$. Since the factorized matrix, \ie $\mathbf{L}$ and $\mathbf{M}$, are the stacked 2D keypoints for all the images, thus $H^1$ and $H^2$ obtained from one image must satisfy the orthonormality constraints on other images, hence we use $\Pi_n \in \mathbb{R}^{2 \times 3K}$ (\ie from image $n$) for our derivation.

Let $\hat{\Pi}_n = [\hat{\Pi}_{n}^1, \hat{\Pi}_{n}^2] = \begin{bmatrix} \hat{\pi}_n^{1,1:K}, &\hat{\pi}_n^{1,K+1:3K} \\ \hat{\pi}_n^{2,1:K}, &\hat{\pi}_n^{2,K+1:3K} \end{bmatrix}$, where $\hat{\pi}_n^{1,1:K}, \hat{\pi}_n^{2,1:K} \in \mathbb{R}^{1 \times K}$ are the first $K$ columns of the first and second rows of $\hat{\Pi}_n$, and $\hat{\pi}_n^{1,K+1:3K}, \hat{\pi}_n^{2,K+1:3K} \in \mathbb{R}^{1 \times 2K}$ are the last $2K$ columns of the first and second rows of $\hat{\Pi}_n$, respectively. Thus, Eq. \eqref{V_PF} implies:
\begin{align}
&L_n = \hat{\Pi}^1_n H^1 (H^1)^{-1} \hat{\mathbf{V}}_x =
\begin{bmatrix}
  r^{11}_n \\ r^{21}_n
\end{bmatrix} \mathbf{z}_n \mathbf{V}_x, \label{U_PF2} \\
&M_n = \hat{\Pi}^2_n H^2 (H^2)^{-1} \hat{\mathbf{V}}_{yz} =
\begin{bmatrix}
  r_n^{1,2:3} \\ r_n^{2,2:3}
\end{bmatrix} (\mathbf{z}_n \otimes I_2) \mathbf{V}_{yz}, \label{V_PF2}
\end{align}
where $L_n, M_n \in \mathbb{R}^{2 \times P}$ are the $n$'th double-row of $\mathbf{L}, \mathbf{M}$. $[r_n^{11}, r_n^{12}]^T$ is the first column of the camera projection matrix of the $n$'th image $R_n$, and $[(r_n^{1,2:3})^T, (r_n^{2,2:3})^T]^T$ is the second and third columns of $R_n$.

Let $h_k^1 \in \mathbb{R}^{K \times 1}, h_k^2 \in \mathbb{R}^{2K \times 2}$ be the $k$th column and double-column of $H^1, H^2$, respectively. Then, from Eqs. \eqref{U_PF2} and \eqref{V_PF2}, we get:
\begin{equation}
\hat{\Pi}_n^1 h_k^1 =
\begin{bmatrix}
\hat{\pi}_n^{1,1K} \\
\hat{\pi}_n^{2,1K}
\end{bmatrix}
h_k^1 =
z_{nk}
\begin{bmatrix}
r_n^{11} \\
r_n^{21}
\end{bmatrix}
\qquad
\hat{\Pi}_n^2 h_k^2 =
\begin{bmatrix}
\hat{\pi}_n^{1,K+1:3K} \\
\hat{\pi}_n^{2,K+1:3K}
\end{bmatrix}
h_k^2 =
z_{nk}
\begin{bmatrix}
r_n^{1,2:3} \\
r_n^{2,2:3}
\end{bmatrix} \label{pi_2}
\end{equation}

By merging the equations of Eq. \eqref{pi_2} together, $R_n$ can be represented by:
\begin{equation}
[\hat{\Pi}_n^1 h_k^1, \hat{\Pi}_n^2 h_k^2]
=\begin{bmatrix}
\hat{\pi}_n^{1,1:K}, &\hat{\pi}_n^{1,K+1:3K} \\
\hat{\pi}_n^{2,1:K}, &\hat{\pi}_n^{2,K+1:3K}
\end{bmatrix}
\begin{bmatrix}
h_k^1, & \mathbf{0}_{K \times 2K} \\
\mathbf{0}_{2K \times K}, & h_k^2
\end{bmatrix}
= z_{nk}R_n.
\label{PF_R}
\end{equation}

\begin{remark}
Similar to the rigid symmetry case in \cite{Gao2016_Rigid}, Eq. \eqref{PF_R} indicates that there is no rotation ambiguities on the symmetric direction. The rotation ambiguities only exist in the $yz$-plane (\ie the non-symmetric plane). \label{Remark3}
\end{remark}

The orthonormality constraints $R_n R_n^T = I$ can be imposed to estimate $h_k^1, h_k^2$:
\begin{align}
&[\hat{\Pi}_n^1 h_k^1, \hat{\Pi}_n^2 h_k^2] [\hat{\Pi}_n^1 h_k^1, \hat{\Pi}_n^2 h_k^2]^T = z_{nk}^2 I \nonumber \\
=&
\begin{bmatrix}
\hat{\pi}_n^{1,1:K}, &\hat{\pi}_n^{1,K+1:3K} \\
\hat{\pi}_n^{2,1:K}, &\hat{\pi}_n^{2,K+1:3K}
\end{bmatrix}
\begin{bmatrix}
h_k^1 h_k^{1T}, & \mathbf{0}_{K \times 2} \\
\mathbf{0}_{2K \times 1}, & h_k^2 h_k^{2T}
\end{bmatrix}
\begin{bmatrix}
\hat{\pi}_n^{1,1:K}, &\hat{\pi}_n^{1,K+1:3K} \\
\hat{\pi}_n^{2,1:K}, &\hat{\pi}_n^{2,K+1:3K}
\end{bmatrix}^T
\end{align}
Thus, we have:
\begin{align}
&\hat{\pi}_n^{1,1:K} h_k^1 h_k^{1T} (\hat{\pi}_n^{1,1:K})^T +  \hat{\pi}_n^{1,K+1:3K} h_k^2 h_k^{2T} (\hat{\pi}_n^{1,K+1:3K})^T = z_{nk}^2 \label{NR_orth1}\\
&\hat{\pi}_n^{2,1:K} h_k^1 h_k^{1T} (\hat{\pi}_n^{2,1:K})^T +  \hat{\pi}_n^{2,K+1:3K} h_k^2 h_k^{2T} (\hat{\pi}_n^{2,K+1:3K})^T = z_{nk}^2 \label{NR_orth2}\\
&\hat{\pi}_n^{1,1:K} h_k^1 h_k^{1T} (\hat{\pi}_n^{2,1:K})^T +  \hat{\pi}_n^{1,K+1:3K} h_k^2 h_k^{2T} (\hat{\pi}_n^{2,K+1:3K})^T = 0
\end{align}

\begin{remark}
The main difference of the derivations from the orthonormality constraints between the rigid and non-rigid cases is that, for the rigid case, the dot product of each row of $\mathbf{R}$ is equal to 1, while for non-rigid the dot product on each row of $\mathbf{\Pi}$ gives us a unknown value $z_{nk}^2$. But note that $z_{nk}^2$ is the same for the both rows, \ie Eqs. \eqref{NR_orth1} and \eqref{NR_orth2}, corresponding to the same projection.
\end{remark}

Eliminating the unknown value $z_{nk}^2$ in Eqs. \eqref{NR_orth1} and \eqref{NR_orth2} (by subtraction) and rewriting in vectorized form gives:
{\small
\begin{align}
& \begin{bmatrix}
\hat{\pi}_n^{1,1:K} \otimes \hat{\pi}_n^{1,1:K} - \hat{\pi}_n^{2,1:K} \otimes \hat{\pi}_n^{2,1:K}, & \hat{\pi}_n^{1,K+1:3K} \otimes \hat{\pi}_n^{1,K+1:3K} - \hat{\pi}_n^{2,K+1:3K} \otimes \hat{\pi}_n^{1,K+1:3K} \\
\hat{\pi}_n^{1,1:K} \otimes \hat{\pi}_n^{2,1:K}, & \hat{\pi}_n^{1,K+1:3K} \otimes \hat{\pi}_n^{2,K+1:3K}
\end{bmatrix} \nonumber \\
& \cdot \begin{bmatrix}
\v(h_k^1 h_k^{1T}) \\
\v(h_k^2 h_k^{2T})
\end{bmatrix}
 =  A_n \begin{bmatrix}
\v(h_k^1 h_k^{1T}) \\
\v(h_k^2 h_k^{2T})
\end{bmatrix}
 = 0,
\end{align}
}

Letting $\mathbf{A} = [A_1^T, ..., A_N^T]^T$, yield the constraints:
\begin{equation}
\mathbf{A} [\v(h_k^1 h_k^{1T})^T, \v(h_k^2 h_k^{2T})^T]^T = 0.  \label{PFconstrain}
\end{equation}

\begin{remark}
As shown in Xiao \etal \cite{Xiao04}, the orthonormality constraints, \ie Eq. \eqref{PFconstrain}, are not sufficient to solve for the ambiguity matrix $H$. But Xiao \etal showed that the solution of $[\v(h_k^1 h_k^{1T})^T, \v(h_k^2 h_k^{2T})^T]^T$ lies in the null space of $\mathbf{A}$ of dimensionality $(2K^2 - K)$ \cite{Xiao04}. Akhter \etal \cite{Akhter2011} proved that this was a ``gauge freedom'' because all legitimate solutions lying in this subspace (despite under-constrained) gave the same solutions for the 3D structure. More technically, the ambiguity of  $H$ corresponds only to a linear combination of $H$'s column-triplet and a rotation on $H$ \cite{Akhter09}. This observation was exploited by Dai \etal in \cite{Dai12,Dai14}, where they showed that, up to the ambiguities aforementioned, $h_k h_k^T$ can be solved by the intersection of 3 subspaces  as we will describe in the following.
\end{remark}

Following the strategy in \cite{Dai12,Dai14}, we have intersection of  subspaces conditions:
\begin{equation}
\left \{ \mathbf{A}\begin{bmatrix}
\v(h_k^1 h_k^{1T}) \\
\v(h_k^2 h_k^{2T})
\end{bmatrix} = 0\right\} \cap \left \{ \begin{matrix} h_k^1 h_k^{1T} \succeq 0 \\ h_k^2 h_k^{2T} \succeq 0  \end{matrix}\right\} \cap \left \{\begin{matrix} \text{rank}(h_k^1 h_k^{1T}) =1 \\ \text{rank}(h_k^2 h_k^{2T}) =2 \end{matrix} \right\} \label{subspaces}
\end{equation}

The first subspace comes from Eq. \eqref{PFconstrain}, \ie the solutions of the Eq. \eqref{PFconstrain} lie in the the null space of $\mathbf{A}$ of dimensionality $(2K^2 - K)$ \cite{Xiao04}. The second subspace requires that $h_k^1 h_k^{1T}$ and $h_k^2 h_k^{2T}$ are positive semi-definite. The third subspace comes from the fact that $h_k^1$ is of rank 1 and $h_k^2$ is of rank 2.

Note that as stated in \cite{Dai12,Dai14}, Eq. \eqref{subspaces} imposes all the necessary constraints on $[\v(h_k^1 $ $ h_k^{1T})^T, \v(h_k^2 h_k^{2T})^T]^T$. There is no difference in the recovered 3D structures using the different solutions that satisfy Eq. \eqref{subspaces}.

We can obtain a solution of $[\v(h_k^1 h_k^{1T})^T, \v(h_k^2 h_k^{2T})^T]^T$, under the condition of Eq. \eqref{subspaces}, by standard semi-definite programming (SDP):
\begin{align}
&\min ||h_k^1 h_k^{1T}||_{\ast} + ||h_k^2 h_k^{2T}||_{\ast} \nonumber \\
 \text{s. t. }  h_k^1 h_k^{1T} \succeq 0, \quad & h_k^2 h_k^{2T} \succeq 0 \quad \mathbf{A} [\v(h_k^1 h_k^{1T})^T, \v(h_k^2 h_k^{2T})^T]^T = 0,
\end{align}
where $|| \cdot ||_{\ast}$ indicates the trace norm.

\begin{remark}
After recovering $h_k^1$ and $h_k^2$, we can estimate the camera parameters $R$ as follows. Note that it does not need to the whole ambiguity matrix $H$ \cite{Dai12,Dai14}.
\end{remark}

After $h_k^1, h_k^2$ has been solved, Eq. \eqref{PF_R} (\ie $[\hat{\Pi}_{n}^1 h_k^1, \hat{\Pi}_{n}^2 h_k^2] = z_{nk}R_n$) implies that the camera projection matrix $R_n$ can be obtained by normalizing the two rows of $[\hat{\Pi}_{n}^1 h_k^1, \hat{\Pi}_{n}^2 h_k^2]$ to have unit $\ell_2$ norm. Then, $\mathbf{R}$ can be constructed by $\mathbf{R} = \text{blkdiag}([R_1, ..., R_N])$.

\begin{remark}
After estimated the camera parameters, we can solve for the 3D structure adopting the methods in \cite{Dai12,Dai14}, \ie by minimizing a \emph{low-rank} constraint on rearranged (\ie more compact) $\mathbf{S}^{\sharp}$ under the orthographic projection model.
\end{remark}

Similar to \cite{Dai12,Dai14}, the structure $\mathbf{S}$ can be estimated by:
\begin{align}
&\min ||\mathbf{S}^{\sharp}||_{\ast} \nonumber \\
\text{s. t. }  [\mathbf{Y} ,\mathbf{Y}^{\dag}] = \mathbf{R} [\mathbf{S}, \a_N & \mathbf{S}] \qquad \mathbf{S}^{\sharp} = [\mathcal{P}_x, \mathcal{P}_y, \mathcal{P}_z] (I_3 \otimes \mathbf{S}),  \label{PF_S}
\end{align}
where $\mathcal{A}_N = I_N \otimes \text{diag}([-1, 1, 1])$, $\mathbf{S} = [S_1^T, ..., S_N^T]^T \in \mathbb{R}^{3N \times P}$ and $\mathbf{S}^{\sharp} \in \mathbb{R}^{N \times 3P}$ is rearranged and more compact $\mathbf{S}$, \ie
\begin{equation}
\mathbf{S}^{\sharp} =
\begin{bmatrix}
x_{11}, &..., &x_{1P}, &y_{11}, &..., &y_{1P}, &z_{11}, &..., & z_{1P} \\
\vdots & & \vdots & \vdots & & \vdots & \vdots & & \vdots \\
x_{N1}, &..., &x_{NP}, &y_{N1}, &..., &y_{NP}, &z_{N1}, &..., & z_{NP}
\end{bmatrix} \nonumber,
\end{equation}
and $\mathcal{P}_x, \mathcal{P}_y, \mathcal{P}_z \in \mathbb{R}^{N \times 3N}$ are the row-permutation matrices of 0 and 1 that select $(I_3 \otimes \mathbf{S})$ to form $\mathbf{S}^{\sharp}$, \ie $\mathcal{P}_x(i, 3i-2) = 1, \mathcal{P}_y(i, 3i-1) = 1, \mathcal{P}_z(i, 3i) = 1$ for $i = 1,...,N$.

\begin{remark}
After obtaining the initial estimates of $R_n, S_n$ (from matrix factorization as described above) and the occluded keypoints, we can minimize the full energy (likelihood) in Eq. \eqref{PF1} d by \emph{coordinate descent} to obtain better estimates of $R_n, S_n$ and the occluded keypoints.
\end{remark}

\emph{Energy Minimization} After obtained initial $\mathbf{R}$, $\mathbf{S}$ and missing points, Eq. \eqref{PF1} can be minimized by coordinate descent. The energy about $\mathbf{R}$, $\mathbf{S}$ is:
\begin{equation}
\mathcal{Q}(\mathbf{R}, \mathbf{S}) =  ||\mathbf{Y} - \mathbf{R} \mathbf{S}||^2_2 + ||\mathbf{Y}^{\dag} - \mathbf{R} \a_K \mathbf{S}||^2_2, \label{PFNREner}
\end{equation}

Note that $\mathbf{S}$ can be updated exactly as the same as its initialization in Eq. \eqref{PF_S} by the low-rank constraint. While each $R_n$ of $\mathbf{R}$ should be updated under the nonlinear orthonormality constraints $R_n R_n^T = I$ similar to the idea in EM-PPCA \cite{Torresani08}: we first parameterize $R_n$ to a full $3 \times 3$ rotation matrix $Q$ and update $Q$ by its rotation increment. Please refer to Appendix \ref{sect:R}.

The occluded points $Y_{n,p}$ and $Y_{n,p}^{\dag}$ with $(n,p) \in IVS$ are updated by minimizing the full energy in Eq. \eqref{PF1} directly:
\begin{equation}
Y_{n,p} = R_n S_p, \qquad Y_{n,p}^{\dag} = R_n \a S_{n,p}
\end{equation}

Similar to  Sym-RSfM \cite{Gao2016_Rigid}, after updating the occluded points, we also re-estimate the translation for each image by $t_n = \sum_p (Y_{n,p} - R_n S_p + Y_{n,p}^{\dag} - R_n \mathcal{A} S_p)$, then centralize the data again by $Y_n \gets Y_n - \mathbf{1}_P^T \otimes t_n$ and $Y_n^{\dag} \gets Y_n^{\dag} - \mathbf{1}_P^T \otimes t_n$.

\section{Experiments}
\subsection{Experimental Settings}
We follow the experimental settings in \cite{Kar15}, using the 2D annotations in \cite{Bourdev10} and 3D keypoints in Pascal3D+ \cite{Xiang14}. Although Pascal3D+ is the best 3D dataset available, it still has some limitations for our task. Specifically, it does not have the complete 3D models for each object; instead it provides the 3D shapes of object subtypes. For example, it provides 10 subtypes for \emph{car} category, such as \emph{sedan, truck}, which ignores the shape variance within each subtype.

Similar to \cite{Dai12,Dai14,Akhter08,Gotardo11}, the rotation error $e_R$ and shape error $e_S$ are used for evaluation. We normalize 3D groundtruth and our 3D estimates to eliminate different scales they may have. For each shape $S_n$ we use its standard deviations in $X, Y, Z$ coordinates $\sigma_n^x, \sigma_n^y, \sigma_n^z$ for the normalization: $S_n^{\text{norm}} =  3 S_n/(\sigma_n^x + \sigma_n^y + \sigma_n^z)$. To deal with the rotation ambiguity between the 3D groundtruth and our reconstruction, we use the Procrustes method \cite{Schonemann66}  to align them. Then, the the rotation error $e_R$ and shape error $e_S$ can be calculated as:
\begin{equation}
e_R = \frac{1}{N} \sum_{n=1}^{N} ||R_n^{\text{aligned}} - R_n^* ||_F, \ \
e_S = \frac{1}{2NP} \sum_{n=1}^{N} \sum_{p = 1}^{2P} ||S_{n,p}^{\text{norm aligned}} - S_{n,p}^{\text{norm}*}||_F,
\label{eq:Error}
\end{equation}
where $R_n^{\text{aligned}}$ and $R_n^*$ are the recovered and the groundtruth camera projection matrix for image $n$. $S_{n,p}^{\text{norm aligned}}$ and $S_{n,p}^{\text{norm}*}$ are the normalized estimated and the normalized groundtruth structure for the $p$'th point of image $n$. $R_n^{\text{aligned}}$ and $R_n^*$, $S_{n,p}^{\text{norm aligned}}$ and $S_{n,p}^{\text{norm}*}$ are aligned by Procrustes method \cite{Schonemann66}.

\begin{table*}[t!]
\setlength{\belowcaptionskip}{-15pt}
\fontsize{8pt}{0.9\baselineskip}\selectfont
\begin{tabular*}{1.006\textwidth}{@{\extracolsep{\fill}} || l || c | c | c | c | c | c | c || c || c | c | c | c | c | c || c ||}
\hline
& \multicolumn{8}{c||}{\textbf{aeroplane}} & \multicolumn{7}{c||}{\textbf{bus}} \\
\hline
& I & II & III & IV & V & VI & VII & {\scriptsize mRE} & I & II & III & IV & V & VI & {\scriptsize mRE} \\
\hline
EP   & 0.36 & 0.59 & 0.50 & 0.49 & 0.57 & 0.57 & \underline{\textbf{0.45}} & 0.34 & 0.42 & 0.34 & 0.56 & 0.54 & 0.98 & 0.86 & 0.26 \\
PF   & 0.99 & 1.08 & 1.13 & 1.15 & 1.22 & 1.10 & 1.11 & 0.52 & 1.62 & 1.56 & 1.75 & 1.59 & 2.09 & 1.70 & 0.47\\
Sym-EP       & \underline{\textbf{0.33}} & \underline{\textbf{0.53}} & \underline{\textbf{0.46}} & \underline{\textbf{0.43}} & \underline{\textbf{0.51}} & \underline{\textbf{0.53}} & 0.46 & \underline{\textbf{0.31}} &\underline{\textbf{0.28}} & \underline{\textbf{0.25}} & \underline{\textbf{0.33}} & \underline{\textbf{0.33}} & \underline{\textbf{0.65}} & \underline{\textbf{0.46}} & \underline{\textbf{0.21}} \\
Sym-PF      & 0.57 & 0.76 & 0.84 & 0.76 & 0.73 & 0.61 & 0.79 & 0.46 & 1.92 & 1.95 & 1.77 & 1.54 & 1.70 & 1.42 & 1.23\\
\hline
\end{tabular*}

\begin{tabular*}{1.006\textwidth}{@{\extracolsep{\fill}} || l || c | c | c | c | c | c | c | c | c | c || c || c | c | c ||}
\hline
& \multicolumn{11}{c||}{\textbf{car}} & \multicolumn{3}{c||}{\textbf{sofa}} \\
\hline
& I & II & III & IV & V & VI & VII & VIII & IX & X & {\scriptsize mRE}  & I & II & III  \\
\hline
EP   & 1.10 & 1.01 & 1.09 & 1.05 & 1.03 & 1.07 & 0.99 & 1.46 & 1.00 & 0.85 & 0.39 & 2.00 & 1.87 & 2.03 \\
PF   & 1.76 & 1.67 & 1.76 & 1.77 & 1.65 & 1.79 & 1.67 & 1.57 & 1.70 & 1.42 & 0.86 & 1.71 & 1.41 & 1.46 \\
Sym-EP       & \underline{\textbf{0.99}} & \underline{\textbf{0.89}} & \underline{\textbf{1.05}} & \underline{\textbf{1.02}} & \underline{\textbf{0.92}} & \underline{\textbf{1.00}} & \underline{\textbf{0.89}} & \underline{\textbf{1.39}} & \underline{\textbf{0.95}} & \underline{\textbf{0.68}} & \underline{\textbf{0.34}} &  \underline{\textbf{1.18}} & \underline{\textbf{0.81}}  & \underline{\textbf{1.08}} \\
Sym-PF      &   1.74 & 1.41 & 1.70 & 1.48 & 1.69 & 1.58 & 1.43 & 1.69 & 1.52 & 1.30 & 0.79 & 1.33 & 1.15 & 1.36 \\
\hline
\end{tabular*}

\begin{tabular*}{1.006\textwidth}{@{\extracolsep{\fill}} || l || c | c | c || c || c | c | c | c || c || c | c | c | c || c ||}
\hline
& \multicolumn{4}{c||}{\textbf{sofa}} & \multicolumn{5}{c||}{\textbf{train}} & \multicolumn{5}{c||}{\textbf{tv}} \\
\hline
 & IV & V & VI & {\scriptsize mRE} & I & II & III & IV & {\scriptsize mRE} & I & II & III & IV & {\scriptsize mRE}  \\
\hline
EP   & 1.99 & 2.37 & 1.81 & 0.79 & 1.18 & 0.53 & 0.49 & 0.42 & 0.85 & \underline{\textbf{0.44}} & \underline{\textbf{0.51}} & \underline{\textbf{0.44}} & \underline{\textbf{0.36}} & \underline{\textbf{0.41}} \\
PF   & 2.02 & 2.66 & 1.64 & 1.36 & 1.97 & \underline{\textbf{0.27}} & 0.47 & 0.34 &  0.98 & 0.56 & 1.01 & 0.97 & 0.65 & 0.80 \\
Sym-EP       & 1.12 & 1.80 & \underline{\textbf{0.88}} & \underline{\textbf{0.34}} & \underline{\textbf{0.95}} & 0.46 & \underline{\textbf{0.42}} & \underline{\textbf{0.31}} & \underline{\textbf{0.73}} &  0.51 & 0.60 & 0.53 & 0.64 & 0.52 \\
Sym-PF       & \underline{\textbf{1.02}} & \underline{\textbf{1.17}} & 0.95 & 0.85 & 1.52 & 0.40 & 0.49 & 0.47  & 0.99 & 0.60 & 1.01 & 1.15 & 0.51 & 0.86 \\
\hline
\end{tabular*}
\caption{The mean \emph{shape} and \emph{rotation} errors for \emph{aeroplane, bus, car, sofa, train, tv}.
% Since Pascal3D+ \cite{Xiang14} provides one 3D model for each subtype, thus the mean shape errors are reported according to each subtype. While the camera projection matrices are available for all the images, thus we report the mean rotation errors for each category.
The Roman numerals indicates the index of subtypes for the mean shape error, and mRE is short for the mean rotation error. EP, PF, Sym-EP, Sym-PF are short for EM-PPCA \cite{Torresani08}, PriorFree \cite{Dai12,Dai14}, Sym-EM-PPCA, Sym-PriorFree, respectively.}
\label{table:Results}
\end{table*}

\subsection{Experimental Results on the Annotated Dataset}
In this section, we construct the 3D keypoints for each image using the non-rigid model. Firstly, we follow the experimental setting in \cite{Kar15} and collect all  images with more than 5 visible keypoints. Then, we do  10 iterations with rank 3 recovery to initialize the occluded/missing data. In this experiments, we use 3 deformation bases and set $\lambda$ in Sym-EM-PPCA, \ie in Eqs. \eqref{likelihood}, as 1.

We tested our algorithm on Pascal \emph{aeroplane, bus, car, sofa, train, tv} based on the mean shape and rotation errors as in Eq. \eqref{eq:Error}. For the shape error, since Pascal3D+ \cite{Xiang14} only provides one 3D model for each subtype, we compare the reconstructed 3D keypoints for each image with their subtype groundtruth. More specifically, the reconstructed 3D keypoints for each image are grouped into subtypes, and we count the mean shape error (we also have median errors in Appendix \ref{sect:median}) by comparing all the images within that subtype to the subtype groundtruth from Pascal3D+. While such problem does not exist for the rotation errors, \ie the groundtruth projection is available for each image in Pascal3D+ \cite{Xiang14}, thus the rotation errors are reported by each category.

The results are reported in Table \ref{table:Results}. The results show that our method outperforms the baselines in general. But we note that our method is not as good as the baselines in some cases, especially for \emph{tv}. The reasons might be: (i) the orthographic projection is inaccurate when the object is close to the camera. Although all the methods used the same suboptimal orthographic projection for these cases, it may deteriorate more on our model sometimes, since we model more constraints. (ii) It might be because the 3D groundtruth in Pascal3D+, which neglects the shape variations in the same subtype, may not be accurate enough  (\eg it has only one 3D model for all the \emph{sedan} cars).

\begin{table*}[t!]
\centering
\fontsize{8pt}{0.9\baselineskip}\selectfont
\setlength{\abovecaptionskip}{5pt}
\setlength{\belowcaptionskip}{-15pt}
\begin{tabular*}{0.95\textwidth}{@{\extracolsep{\fill}} || l || c | c | c | c | c | c | c || c || c | c | c | c ||}
\hline
& \multicolumn{8}{c||}{$\sigma$ = \textbf{0.03} $d_{max}$} & \multicolumn{4}{c||}{$\sigma$ = \textbf{0.05} $d_{max}$} \\
\hline
& I & II & III & IV & V & VI & VII & mRE & I & II & III & IV  \\
\hline
EP  & 0.34 & 0.59 & 0.49 & 0.45 & 0.54 & 0.55 & \underline{\textbf{0.45}} & 0.33 & 0.37 & 0.58 & 0.51 & 0.47 \\
\hline
PF  & 0.92 & 1.01 & 1.05 & 1.06 & 1.13 & 1.03 & 1.06 & 0.52 & 0.93 & 1.04 & 1.05 & 1.08 \\
\hline
Sym-EP &  \underline{\textbf{0.34}} & \underline{\textbf{0.54}} & \underline{\textbf{0.47}} & \underline{\textbf{0.44}} & \underline{\textbf{0.52}} & \underline{\textbf{0.55}} & 0.46 & \underline{\textbf{0.32}} & \underline{\textbf{0.35}} & \underline{\textbf{0.54}} & \underline{\textbf{0.47}} & \underline{\textbf{0.43}} \\
\hline
Sym-PF &   0.79 & 0.93 & 1.01 & 0.93 & 0.91 & 0.79 & 0.94 & 0.60 & 0.83 & 0.99 & 1.09 & 0.98 \\
\hline
\end{tabular*}

\begin{tabular*}{0.95\textwidth}{@{\extracolsep{\fill}} || l || c | c | c || c || c | c | c | c | c | c | c || c ||}
\hline
& \multicolumn{4}{c||}{$\sigma$ = \textbf{0.05} $d_{max}$} & \multicolumn{8}{c||}{$\sigma$ = \textbf{0.07} $d_{max}$} \\
\hline
& V & VI & VII & mRE & I & II & III & IV & V & VI & VII & mRE  \\
\hline
EP  & 0.57 & 0.57 & 0.47 & 0.35 & 0.38 & 0.61 & 0.50 & 0.45 & 0.61 & 0.57 & \underline{\textbf{0.47}} & 0.36 \\
\hline
PF   & 1.15 & 1.02 & 1.07 & 0.54 & 0.94 & 1.04 & 1.08 & 1.07 & 1.15 & 1.03 & 1.08 & 0.65 \\
\hline
Sym-EP   & \underline{\textbf{0.52}} & \underline{\textbf{0.57}} & \underline{\textbf{0.46}} & \underline{\textbf{0.33}} & \underline{\textbf{0.37}} & \underline{\textbf{0.58}} & \underline{\textbf{0.49}} & \underline{\textbf{0.44}} & \underline{\textbf{0.58}} & \underline{\textbf{0.57}} & 0.47 & \underline{\textbf{0.35}} \\
\hline
Sym-PF   & 0.94 & 0.84 & 1.04 & 0.63 & 0.94 & 1.06 & 1.15 & 1.04 & 1.05 & 0.89 & 1.08 & 0.70 \\
\hline
\end{tabular*}

\caption{The mean \emph{shape} and \emph{rotation} errors for \emph{aeroplane} with imperfect annotations. The noise is Gaussian $\mathcal{N}(0, \sigma^2)$ with $\sigma = sd_{max}$, where we choose $s = 0.03, 0.05, 0.07$ and $d_{max}$ is the longest distance between all the keypoints (\ie the tip of the nose to the tip of the tail for aeroplane). Other parameters are the same as those in Table \ref{table:Results}. Each result value is obtained by averaging 10 repetitions.}
\label{table:Err_noise}
\end{table*}

\subsection{Experimental Results on the Noisy Annotations}
We also investigate what happens if the keypoints are not perfectly annotated. This is important to check because our method depends on keypoint pairs therefore may be sensitive to errors in keypoint location, which will inevitably arise when we use features detectors, \eg deep nets \cite{chen2014articulated}, to detect the keypoints.

To simulate this, we add Gaussian noise $\mathcal{N}(0, \sigma^2)$ to the 2D annotations and re-do the experiments. The standard deviation is set to $\sigma = s d_{max}$, where $d_{max}$ is the longest distance between all the keypoints (\eg for \emph{aeroplane}, it is the distance between the nose tip to the tail tip). We have tested for different $s$ by: 0.03, 0.05, 0.07. The other parameters are the same as the previous section.

The results for \emph{aeroplane} with $s = 0.03, 0.05, 0.07$ are shown in Table \ref{table:Err_noise}. Each result value is obtained by averaging 10 repetitions. The results in Table \ref{table:Err_noise} show that the performances of all the methods decrease in general with the increase in the noise level. Nonetheless, our methods still outperform our counterparts with the noisy annotations (\ie the imperfectly labeled annotations).

\section{Conclusion}

This paper shows that non-rigid SfM can be extended to the important special case where the objects are symmetric, which is frequently possessed by man-made objects \cite{Rosen12,Hong04}. We derive and implement this extension to two popular non-rigid structure from motion algorithms \cite{Torresani08,Dai12,Dai14}, which perform well on the Pascal3D+ dataset when compared to the baseline methods.

In this paper, we have focused on constructing the non-rigid SfM model(s) that can exploit the symmetry property. In future work, we will extend to perspective projection, apply a better initialization of the occluded keypoints such as low-rank recovery, use additional object features (instead of just key-points), and detect these features from images automatically such as \cite{chen2014articulated}.

\section{Acknowledgment}

We would like to thank Ehsan Jahangiri, Cihang Xie, Weichao Qiu, Xuan Dong, Siyuan Qiao for giving feedbacks on the manuscript. This work was supported by ARO 62250-CS and ONR N00014-15-1-2356.

\clearpage

\bibliographystyle{splncs}
\bibliography{egbib2}

\begin{thebibliography}{10}

\bibitem{Tomasi92}
Tomasi, C., Kanade, T.:
\newblock Shape and motion from image streams under orthography: a
  factorization method.
\newblock International Journal of Computer Vision \textbf{9}(2) (1992)
  137--154

\bibitem{Hartley2004}
Hartley, R.I., Zisserman, A.:
\newblock Multiple View Geometry in Computer Vision. Second edn.
\newblock Cambridge University Press (2004)

\bibitem{Torresani03}
Torresani, L., Hertzmann, A., Bregler, C.:
\newblock Learning non-rigid 3d shape from 2d motion.
\newblock In: NIPS. (2003)

\bibitem{Xiao04}
Xiao, J., Chai, J., Kanade, T.:
\newblock A closed-form solution to nonrigid shape and motion recovery.
\newblock In: ECCV. (2004)

\bibitem{Torresani08}
Torresani, L., Hertzmann, A., Bregler, C.:
\newblock Nonrigid structure-from-motion: Estimating shape and motion with
  hierarchical priors.
\newblock IEEE Transactions on Pattern Analysis and Machine Intelligence
  \textbf{30} (2008)  878--892

\bibitem{Akhter2011}
Akhter, I., Sheikh, Y., Khan, S., Kanade, T.:
\newblock Trajectory space: A dual representation for nonrigid structure from
  motion.
\newblock IEEE Transactions on Pattern Analysis and Machine Intelligence
  \textbf{33}(7) (2011)  1442--1456

\bibitem{Gotardo11}
Gotardo, P., Martinez, A.:
\newblock Computing smooth timetrajectories for camera and deformable shape in
  structure from motion with occlusion.
\newblock IEEE Transactions on Pattern Analysis and Machine Intelligence
  \textbf{33} (2011)  2051--2065

\bibitem{Hamsici2012}
Hamsici, O.C., Gotardo, P.F., Martinez, A.M.:
\newblock Learning spatially-smooth mappings in non-rigid structure from
  motion.
\newblock In: ECCV. (2012)  260--273

\bibitem{Dai12}
Dai, Y., Li, H., He, M.:
\newblock A simple prior-free method for non-rigid structure-from-motion
  factorization.
\newblock In: CVPR. (2012)

\bibitem{Dai14}
Dai, Y., Li, H., He, M.:
\newblock A simple prior-free method for non-rigid structure-from-motion
  factorization.
\newblock International Journal of Computer Vision \textbf{107} (2014)
  101--122

\bibitem{ma2015non}
Ma, J., Zhao, J., Ma, Y., Tian, J.:
\newblock Non-rigid visible and infrared face registration via regularized
  gaussian fields criterion.
\newblock Pattern Recognition \textbf{48}(3) (2015)  772--784

\bibitem{ma2013robust}
Ma, J., Zhao, J., Tian, J., Tu, Z., Yuille, A.L.:
\newblock Robust estimation of nonrigid transformation for point set
  registration.
\newblock In: CVPR. (2013)

\bibitem{ma2013regularized}
Ma, J., Zhao, J., Tian, J., Bai, X., Tu, Z.:
\newblock Regularized vector field learning with sparse approximation for
  mismatch removal.
\newblock Pattern Recognition \textbf{46}(12) (2013)  3519--3532

\bibitem{Agudo14}
Agudo, A., Agapito, L., Calvo, B., Montiel, J.:
\newblock Good vibrations: A modal analysis approach for sequential non-rigid
  structure from motion.
\newblock In: CVPR. (2014)  1558--1565

\bibitem{Rosen12}
Rosen, J.:
\newblock Symmetry discovered: Concepts and applications in nature and science.
\newblock Dover Publications (2011)

\bibitem{Hong04}
Hong, W., Yang, A.Y., Huang, K., Ma, Y.:
\newblock On symmetry and multiple-view geometry: Structure, pose, and
  calibration from a single image.
\newblock International Journal of Computer Vision \textbf{60} (2004)
  241­--265

\bibitem{Gordon90}
Gordon, G.G.:
\newblock Shape from symmetry.
\newblock In: Proc. SPIE. (1990)

\bibitem{kontsevich93}
Kontsevich, L.L.:
\newblock Pairwise comparison technique: a simple solution for depth
  reconstruction.
\newblock JOSA A \textbf{10}(6) (1993)  1129--1135

\bibitem{Vetter94}
Vetter, T., Poggio, T.:
\newblock Symmetric 3d objects are an easy case for 2d object recognition.
\newblock Spatial Vision \textbf{8} (1994)  443--453

\bibitem{Mukherjee95}
Mukherjee, D.P., Zisserman, A., Brady, M.:
\newblock Shape from symmetry: Detecting and exploiting symmetry in affine
  images.
\newblock Philosophical Transactions: Physical Sciences and Engineering
  \textbf{351} (1995)  77--106

\bibitem{Thrun05}
Thrun, S., Wegbreit, B.:
\newblock Shape from symmetry.
\newblock In: ICCV. (2005)

\bibitem{Li07}
Li, Y., Pizlo, Z.:
\newblock Reconstruction of shapes of 3d symmetric objects by using planarity
  and compactness constraints.
\newblock In: Proc. of SPIE-IS\&T Electronic Imaging. (2007)

\bibitem{Vicente14}
Vicente, S., Carreira, J., Agapito, L., Batista, J.:
\newblock Reconstructing pascal voc.
\newblock In: CVPR. (2014)

\bibitem{Kar15}
Kar, A., Tulsiani, S., Carreira, J., Malik, J.:
\newblock Category-specific object reconstruction from a single image.
\newblock In: CVPR. (2015)

\bibitem{Akhter09}
Akhter, I., Sheikh, Y., Khan, S.:
\newblock In defense of orthonormality constraints for nonrigid structure from
  motion.
\newblock In: CVPR. (2009)

\bibitem{Gao2016_Rigid}
Gao, Y., Yuille, A.L.:
\newblock Exploiting symmetry and/or {M}anhattan properties for 3{D} object
  structure estimation from single and multiple images.
\newblock arXiv preprint arXiv:1607.07129 (2016)

\bibitem{Xiang14}
Xiang, Y., Mottaghi, R., Savarese, S.:
\newblock Beyond pascal: A benchmark for 3d object detection in the wild.
\newblock In: WACV. (2014)

\bibitem{grossmann2002maximum}
Grossmann, E., Santos-Victor, J.:
\newblock Maximum likehood 3d reconstruction from one or more images under
  geometric constraints.
\newblock In: BMVC. (2002)

\bibitem{grossmann2005least}
Grossmann, E., Santos-Victor, J.:
\newblock Least-squares 3d reconstruction from one or more views and geometric
  clues.
\newblock Computer Vision and Image Understanding \textbf{99}(2) (2005)
  151--174

\bibitem{grossmann2002single}
Grossmann, E., Ortin, D., Santos-Victor, J.:
\newblock Single and multi-view reconstruction of structured scenes.
\newblock In: ACCV. (2002)

\bibitem{kontsevich87}
Kontsevich, L.L., Kontsevich, M.L., Shen, A.K.:
\newblock Two algorithms for reconstructing shapes.
\newblock Optoelectronics, Instrumentation and Data Processing \textbf{5}
  (1987)  76--81

\bibitem{Bregler00}
Bregler, C., Hertzmann, A., Biermann, H.:
\newblock Recovering non-rigid 3d shape from image streams.
\newblock In: CVPR. (2000)

\bibitem{hongsecrets}
Hong, J.H., Fitzgibbon, A.:
\newblock Secrets of matrix factorization: Approximations, numerics, manifold
  optimization and random restarts.
\newblock In: ICCV. (2015)

\bibitem{Olsen08}
Olsen, S.I., Bartoli, A.:
\newblock Implicit non-rigid structure-from-motion with priors.
\newblock Journal of Mathematical Imaging and Vision \textbf{31}(2-3) (2008)
  233--244

\bibitem{Akhter08}
Akhter, I., Sheikh, Y., Khan, S., Kanade, T.:
\newblock Nonrigid structure from motion in trajectory space.
\newblock In: NIPS. (2008)

\bibitem{Ceylan14}
Ceylan, D., Mitra, N.J., Zheng, Y., Pauly, M.:
\newblock Coupled structure-from-motion and 3d symmetry detection for urban
  facades.
\newblock ACM Transactions on Graphics \textbf{33} (2014)

\bibitem{morris2001gauge}
Morris, D.D., Kanatani, K., Kanade, T.:
\newblock Gauge fixing for accurate 3d estimation.
\newblock In: CVPR. (2001)

\bibitem{Bishop06}
Bishop, C.M.:
\newblock Pattern Recognition and Machine Learning.
\newblock Springer, New York (2006)

\bibitem{Bourdev10}
Bourdev, L., Maji, S., Brox, T., Malik, J.:
\newblock Detecting people using mutually consistent poselet activations.
\newblock In: ECCV. (2010)

\bibitem{Schonemann66}
Sch{\"o}nemann, P.H.:
\newblock A generalized solution of the orthogonal procrustes problem.
\newblock Psychometrika \textbf{31} (1966)  1--10

\bibitem{chen2014articulated}
Chen, X., Yuille, A.L.:
\newblock Articulated pose estimation by a graphical model with image dependent
  pairwise relations.
\newblock In: NIPS. (2014)  1736--1744

\end{thebibliography}

\newpage
\renewcommand\thesubsection{A\arabic{subsection}}
\renewcommand\thetable{A\arabic{table}}
\section*{Appendix \label{sect:app}}

This appendix contains more details of \emph{Symmetric Non-Rigid Structure from Motion for Category-Specific Object Structure Estimation}:
\begin{enumerate}
	\item In Sect. \ref{sect:median}, we show our results using \emph{median} shape and rotation errors.
	\item In Sect. \ref{sect:M}, we give the update of each parameters in \textbf{M-step} of \emph{Sym-EM-PPCA}.
	\item In Sect. \ref{sect:R}, the minimization of the energy function of \emph{Sym-PriorFree} \wrt the camera projection matrix $R_n$ under orthogonality constrains is detailed.
	% \item Finally, our companion paper, namely Exploiting Symmetry and/or Manhattan Properties for 3D Object Structure Estimation from Single and Multiple Images (Paper ID: 370), is attached.
\end{enumerate}

\subsection{Experimental Results using Median Shape and Rotation Errors \label{sect:median}}
In this section, the \textbf{median} \emph{shape} and \emph{rotation} errors are reported. We used the same parameters as them in the main text, \ie we use 3 deformation bases and set $\lambda$ in Sym-EM-PPCA as 1. The same conclusions can be made by the median errors as them made by the mean errors in the main text.

\begin{table*}[h]
\setlength{\belowcaptionskip}{-10pt}
\fontsize{8pt}{0.9\baselineskip}\selectfont
\begin{tabular*}{1.006\textwidth}{@{\extracolsep{\fill}} || l || c | c | c | c | c | c | c || c || c | c | c | c | c | c || c ||}
\hline
& \multicolumn{8}{c||}{\textbf{aeroplane}} & \multicolumn{7}{c||}{\textbf{bus}} \\
\hline
& I & II & III & IV & V & VI & VII & {\scriptsize mRE} & I & II & III & IV & V & VI & {\scriptsize mRE} \\
\hline
EP   &  0.29 & 0.50 & 0.39 & 0.41 & 0.5 & 0.46 & \underline{\textbf{0.38}} & 0.24 & 0.34 & 0.28 & 0.54 & 0.49 & 0.92 & 0.80 & 0.15\\
PF   &  0.94 & 1.14 & 1.21 & 1.16 & 1.13 & 1.05 & 1.17 & 0.41 & 1.50 & 1.32 & 1.79 & 1.53 & 2.36 & 1.56 & 0.33 \\
Sym-EP       &  \underline{\textbf{0.25}} & \underline{\textbf{0.53}} & \underline{\textbf{0.34}} & \underline{\textbf{0.36}} & \underline{\textbf{0.43}} & \underline{\textbf{0.42}} & 0.40 & \underline{\textbf{0.23}} & \underline{\textbf{0.25}} & \underline{\textbf{0.24}} & \underline{\textbf{0.32}} & \underline{\textbf{0.27}} & \underline{\textbf{0.65}} & \underline{\textbf{0.38}} & \underline{\textbf{0.12}} \\
Sym-PF      & 0.47 & 0.70 & 0.79 & 0.52 & 0.60 & 0.46 & 0.65 & 0.33 & 1.95 & 2.07 & 1.75 & 1.52 & 1.46 & 1.01 & 1.34\\
\hline
\end{tabular*}

\begin{tabular*}{1.006\textwidth}{@{\extracolsep{\fill}} || l || c | c | c | c | c | c | c | c | c | c || c || c | c | c ||}
\hline
& \multicolumn{11}{c||}{\textbf{car}} & \multicolumn{3}{c||}{\textbf{sofa}} \\
\hline
& I & II & III & IV & V & VI & VII & VIII & IX & X & {\scriptsize mRE}  & I & II & III  \\
\hline
EP   &  1.05 & 1.01 & 1.07 & 1.01 & 0.99 & 1.07 & 0.94 & 1.44 & 0.95 & 0.84 & 0.37 & 1.99 & 1.86 & 2.01 \\
PF   &  1.71 & 1.66 & 1.73 & 1.79 & 1.60 & 1.72 & 1.75 & 1.48 & 1.70 & 1.36 & 0.81 & 1.76 & 1.55 & 1.32 \\
Sym-EP       &  \underline{\textbf{0.93}} & \underline{\textbf{0.88}} & \underline{\textbf{1.02}} & \underline{\textbf{0.99}} & \underline{\textbf{0.87}} & \underline{\textbf{0.99}} & \underline{\textbf{0.83}} & \underline{\textbf{1.40}} & \underline{\textbf{0.91}} & \underline{\textbf{0.67}} & \underline{\textbf{0.32}} & \underline{\textbf{1.12}} & \underline{\textbf{0.81}} & \underline{\textbf{1.07}} \\
Sym-PF      &  1.73 & 1.26 & 1.81 & 1.43 & 1.62 & 1.54 & 1.32 & 1.70 & 1.47 & 1.16 & 0.67 & 1.14 & 1.08 & 1.18 \\
\hline
\end{tabular*}

\begin{tabular*}{1.006\textwidth}{@{\extracolsep{\fill}} || l || c | c | c || c || c | c | c | c || c || c | c | c | c || c ||}
\hline
& \multicolumn{4}{c||}{\textbf{sofa}} & \multicolumn{5}{c||}{\textbf{train}} & \multicolumn{5}{c||}{\textbf{tv}} \\
\hline
 & IV & V & VI & {\scriptsize mRE} & I & II & III & IV & {\scriptsize mRE} & I & II & III & IV & {\scriptsize mRE}  \\
\hline
EP   &  1.98 & 2.34 & 1.81 & 0.76 & 1.04 & 0.45 & 0.49 & 0.42 & 0.71 & \underline{\textbf{0.36}} & \underline{\textbf{0.40}} & \underline{\textbf{0.40}} & 0.34 & \underline{\textbf{0.26}}  \\
PF   &  2.03 & 2.51 & 1.53 & 1.42 & 1.81 & 0.38 & 0.48 & 0.30 & 0.87 & 0.56 & 1.01 & 0.96 & 0.61 & 0.66 \\
Sym-EP       &  1.11 & 1.79 & \underline{\textbf{0.88}} & \underline{\textbf{0.24}} & \underline{\textbf{0.91}} & 0.44 & \underline{\textbf{0.41}} & \underline{\textbf{0.25}} & \underline{\textbf{0.55}} & 0.53 & 0.52 & 0.47 & 0.60 & 0.28 \\
Sym-PF       &  \underline{\textbf{0.87}} & \underline{\textbf{0.90}} & 0.92 & 0.76 & 1.67 & \underline{\textbf{0.32}} & 0.52 & 0.47 & 0.91 & 0.48 & 0.91 & 0.98 & \underline{\textbf{0.13}} & 0.76 \\
\hline
\end{tabular*}
\caption{The \textbf{median} \emph{shape} and \emph{rotation} errors for \emph{aeroplane, bus, car, sofa, train, tv}. Since Pascal3D+ \cite{Xiang14} provides one 3D model for each subtype, thus the mean shape errors are reported according to each subtype. While the camera projection matrices are available for all the images, thus we report the mean rotation errors for each category. The Roman numerals indicates the index of subtypes for the mean shape error, and mRE is short for the mean rotation error. EP, PF, Sym-EP, Sym-PF are short for EM-PPCA \cite{Torresani08}, PriorFree \cite{Dai12,Dai14}, Sym-EM-PPCA, Sym-PriorFree, respectively.}
\label{table:supp_Results}
\end{table*}

\subsection{M-Step of Sym-EM-PPCA \label{sect:M}}
This step is to maximize the complete (joint) log-likelihood $P(\mathbb{Y}_n, \mathbb{Y}_n^{\dag}, \mathbf{V} | z_n; G_n, $ $\mathbb{\bar{S}}, \mathbf{V}^{\dag}, \mathbb{T})$. The complete log-likelihood $Q(\theta)$ is:
\begin{align}
\mathcal{Q}(\theta) =& - \sum_{n} \ln P(\mathbb{Y}_n, \mathbb{Y}_n^{\dag} | z_n; G_n, \mathbb{\bar{S}}, \mathbf{V}, \mathbf{V}^{\dag}, \mathbb{T}) + \lambda ||\mathbf{V}^{\dag} - \a_P \mathbf{V}||^2\nonumber \\
=& - \sum_{n} \left ( \ln P(\mathbb{Y}_n | z_n; G_n, \mathbb{\bar{S}}, \mathbf{V}, \mathbb{T}) + \ln P(\mathbb{Y}_n^{\dag} | z_n; G_n, \mathbb{\bar{S}}, \mathbf{V}^{\dag}, \mathbb{T}) \right) + \lambda ||\mathbf{V}^{\dag} - \a_P \mathbf{V}||^2\nonumber \\
=& 2PN\ln(2\pi\sigma^2) + \frac{1}{2\sigma^2} \sum_n E_{z_n} ||\mathbb{Y}_n - G_n(\mathbb{\bar{S}} + \mathbf{V}z_n) - \mathbb{T}_n||^2 \nonumber \\
&+ \frac{1}{2\sigma^2}\sum_n E_{z_n} ||\mathbb{Y}_n^{\dag} - G_n(\a_P\mathbb{\bar{S}} + \mathbf{V}^{\dag}z_n) - \mathbb{T}_n||^2 + \lambda||\mathbf{V} - \a_P\mathbf{V}^{\dag} ||^2 \nonumber \\
&\text{s. t.} \qquad R_n R_n^T = I, \label{supp_likelihood}
% &+  \left(  \lambda_1||{\color{red} \mathcal{R}}V_1 + \mathbf{a} {\color{red} \mathcal{R}} V_2 ||^2 +  \lambda_2 ||  \mathbf{b} {\color{red} \mathcal{R}} \bar{S}^{mid} ||^2 + \lambda_3 ||  \mathbf{c} {\color{red} \mathcal{R}} D ||^2 \right)
\end{align}
where $\theta = \{G_n, \mathbb{\bar{S}}, \mathbf{V}, \mathbf{V}^{\dag}, \mathbb{T}_n, \sigma^2 \}$. $\mathbb{Y}_n \in \mathbb{R}^{2P \times 1}, \mathbb{\bar{S}} \in \mathbb{R}^{3P \times 1}$, and $\mathbb{T}_n \in \mathbb{R}^{2P \times 1}$ are the stacked vectors of 2D keypoints, 3D mean structure and translations. $G_n = I_{P} \otimes c_n R_n$, in which $c_n$ is the scale parameter for weak perspective projection, $\mathbf{V} = [\mathbb{V}_1, ..., \mathbb{V}_K] \in \mathbb{R}^{3P \times K}$ is the grouped $K$ deformation bases, $z_n \in \mathbb{R}^{K \times 1}$ is the coefficient of the $K$ bases. $\mathcal{A} = I_{P} \otimes \mathcal{A}$, and $\mathcal{A} = \text{diag}([-1, 1, 1])$ is a matrix operator which negates the first row of its right matrix.

We first update the shape parameters $\mathbb{\bar{S}}, \mathbf{V}, \mathbf{V}^{\dag}$ by maximize the log-likelihood $\mathcal{Q}$. Since these 3 parameters are related to each other in their derivations, thus they should be updated jointly by setting the 3 derivations to 0. According to Eq. \eqref{supp_likelihood}, we have:
\begin{align}
&\begin{bmatrix}
& A^{\ast}, & (B^{\ast})^T, & C^{\ast} \\
& B^{\ast}, & D^{\ast} + 2\lambda\sigma^2I_{3PK}, & -I_{K} \otimes 2\lambda\sigma^2 \a_P \\
& B^{\ast} \a_P, & -I_{K} \otimes 2\lambda\sigma^2 \a_P^T, & D^{\ast} + I_{K} \otimes 2\lambda\sigma^2 \a_P^T \a_P \\
\end{bmatrix}
\begin{bmatrix}
\mathbb{S} \\
\v(\mathbf{V}) \\
\v(\mathbf{V}^{\dag})
\end{bmatrix} \nonumber \\
 = &
\begin{bmatrix}
\v \left( \sum_n G_n^T (\mathbb{Y} - \mathbb{T}_n) + \a_P^T G_n^T (\mathbb{Y}^{\dag} - \mathbb{T}_n) \right) \\
\v \left( \sum_n G_n^T (\mathbb{Y} - \mathbb{T}_n) \mu_n^T \right) \\
\v \left( \sum_n G_n^T (\mathbb{Y}^{\dag} - \mathbb{T}_n) \mu_n^T \right),
\end{bmatrix} , \label{supp_SEMPPCA}
\end{align}%
where we have:
\begin{align}
& A^{\ast} = \sum_n G_n^T G_n + \a_P^T G_n^T G_n \a_P, \qquad B^{\ast} = \sum_n \mu_n \otimes G_n^T G_n \nonumber \\
& C^{\ast} = \sum_n \mu_n^T \otimes \a_P^T G_n^T G_n, \qquad D^{\ast} = \sum_n \phi_n^T \otimes G_n^T G_n.   
\end{align}%

The camera parameters $t_n, c_n, R_n$ and the variance of the noise $\sigma^2$ can be updated similarly as Bregler's method \cite{Torresani08}. We first replace some parameters to make the equation to be homogeneous:
\begin{align}
&\tilde{\mathbf{V}} = [\mathbb{S},\mathbf{V}], \qquad \tilde{\mathbf{V}}^{\dag} = [\a_P\mathbb{S},\mathbf{V}^{\dag}], \nonumber \\
&\tilde{\mu_n} = [1, \mu_n^T]^T, \qquad \tilde{\phi_n} =
\begin{bmatrix}
1 & \mu_n^T \\
\mu_n & \phi_n
\end{bmatrix}
\end{align}
Then the estimations of new $\sigma^2, t_n, c_n$ are:
\begin{align}
\sigma^{2} = &\frac{1}{4PN}\sum_n \left( ||\mathbb{Y}_n - \mathbb{T}_n||^2 + ||\mathbb{Y}_n^{\dag} - \mathbb{T}_n||^2  \nonumber \right. \\
&\left. - 2(\mathbb{Y}_n - \mathbb{T}_n)G_n\tilde{\mathbf{V}}\tilde{\mu_n} - 2(\mathbb{Y}_n^{\dag} - \mathbb{T}_n)G_n\tilde{\mathbf{V}}^{\dag}\tilde{\mu_n} \right. \nonumber \\
&\left. + \text{tr}(\tilde{\mathbf{V}}^T G_n^T G_n \tilde{\mathbf{V}} \tilde{\phi_n}) + \text{tr}(\tilde{\mathbf{V}}^{\dag T} G_n^T G_n \tilde{\mathbf{V}}^{\dag} \tilde{\phi_n}) \right) \\
t_n =& \frac{1}{2P} \sum_{p=1}^{P} (\mathbb{Y}_{n,p} - c_n R_n \tilde{\mathbf{V}}_{p}\tilde{\mu_n} + \mathbb{Y}_{n,p}^{\dag} - c_n R_n \tilde{\mathbf{V}}_{p}^{\dag}\tilde{\mu_n}) \\
c_n =& \frac{\sum_{p=1}^{P} \left( \tilde{\mu_n}^T \tilde{\mathbf{V}}_{p}^{T} R_n^T (\mathbb{Y}_{n,p} - t_n) + \tilde{\mu_n}^T \tilde{\mathbf{V}}_{p}^{\dag T} R_n^T (\mathbb{Y}_{n,p}^{\dag} - t_n)  \right)}{\sum_{p=1}^{P} \text{tr}(\tilde{\mathbf{V}}_{p}^{T} R_n^T R_n \tilde{\mathbf{V}}_{p} \tilde{\phi_n} + \tilde{\mathbf{V}}_{p}^{\dag T} R_n^T R_n \tilde{\mathbf{V}}_{p}^{\dag} \tilde{\phi_n})}
\end{align}

Since $R_n$ is subject to a nonlinear orthonormality constraint and cannot be updated in closed form, we follow an alternative approach used in \cite{Torresani08} to parameterize $R_n$ as a complete 3 $\times$ 3 rotation matrix $Q_n$ and update the incremental rotation on $Q_n$ instead, \ie $Q_n^{new} = e^{\xi} Q_n$.

Here, the first and second rows of $Q_n$ is the same as $R_n$, and the third row of $Q_n$ is obtained by the cross product of its first and second rows. The relationship of $Q_n$ and $R_n$ can be revealed by a matrix operator $\mathcal{M}$:
\begin{equation}
R_n = \mathcal{M} Q_n, \qquad \mathcal{M} = \begin{bmatrix} 1, 0, 0 \\ 0, 1, 0 \end{bmatrix}. \label{supp_para_R}
\end{equation}

% Replacing $R_n$ by $Q_n$, Equation \eqref{RigidEner0} can be rewritten as:
% \begin{equation}
% 	\mathcal{Q}(R_n) = \sum_n ||Y_n - \mathcal{M} Q_n S||^2_2 + \sum_n||Y^{\dag}_n - \mathcal{M} Q_n \a S ||^2_2
% \end{equation}

Note that the incremental rotation $e^{\xi}$ can be further approximated by its first order Taylor Series, \ie $e^{\xi} \approx I + \xi$. Finally, we have: 
\begin{equation}
	R_n^{new}(\xi) = \mathcal{M} (I+\xi)Q_n. \label{supp_R_Q}
\end{equation}

Therefore, setting $\partial\mathcal{Q} / \partial R_n = 0$, then replace $R_n$ by $Q_n$ using Eq. \eqref{supp_R_Q} and vectorize it, we have:
\begin{align}
R_n =& \mathcal{M} e^\xi Q_n \approx \mathcal{M} (I + \xi) Q_n \quad \text{  and  } \quad \text{vec}(\xi) = \alpha^{+}\beta \label{20}\\
\alpha =& \left( c_n^2 \sum_{p=1}^{P} (\tilde{\mathbf{V}}_{p}^T \tilde{\phi_n} \tilde{\mathbf{V}}_{p} + \tilde{\mathbf{V}}_{p}^{\dag T} \tilde{\phi_n} \tilde{\mathbf{V}}_{p}^{\dag})^T Q_n^T \right) \otimes \mathcal{M}; \label{21}\\
\beta =& \v\left( c_n \sum_{p=1}^{P} \left( (\mathbb{Y}_{n,p} - t_n) \tilde{\mu_n}^T \tilde{\mathbf{V}}_{p}^T + (\mathbb{Y}_{n,p}^{\dag} - t_n) \tilde{\mu_n}^T \tilde{\mathbf{V}}_{p}^{\dag T} \right) \right. \nonumber \\& - \left. c_n^2 \mathcal{M} Q_n \sum_{p=1}^{P} (\tilde{\mathbf{V}}_{p}^T \tilde{\phi_n} \tilde{\mathbf{V}}_{p} + \tilde{\mathbf{V}}_{p}^{\dag T} \tilde{\phi_n} \tilde{\mathbf{V}}_{p}^{\dag}) \right) \label{22}
\end{align}
where the subscript $p$ means the $p$th keypoint, $\alpha^+$ means the pseudo inverse matrix of $\alpha$, $\otimes$ denotes Kronecker product.

\subsection{Update Camera Projection Matrix $R_n$ Under Orthogonality Constrains for Sym-PriorFree \label{sect:R}}
In the Sym-PriorFree method, the energy function \wrt $R_n$ is:
\begin{equation}
	Q(R_n) = \sum_n||Y_n - R_n S_n||_2^2 + \sum_n||Y^{\dag}_n - R_n \a S_n||_2^2,
\end{equation}
where $Y_n, Y^{\dag}_n \in \mathbb{R}^{2 \times P}, S_n \in \mathbb{R}^{3 \times P}$ are the 2D symmetric keypoint pairs and the 3D structure of image $n$. $R_n \in \mathbb{R}^{2 \times 3}$ is the camera projection. $\mathcal{A} = \text{diag}([-1, 1, 1])$ is a matrix operator which negates the first row of its right matrix.

We use the same updating procedures as used in the previous section to update $R_n$ for Sym-PriorFree method. Specifically, we firstly parameterize $R_n$ as a $Q_n$ by Eq. \eqref{supp_para_R}. Then, $Q_n$ can be updated by $Q_n^{new} = e^{\xi} Q_n \approx (I + \xi) Q_n$. 

Therefore, setting the derivation of $\partial\mathcal{Q} / \partial R_n = 0$, then replace $R_n$ by $Q_n$ using Eq. \eqref{supp_R_Q} and vectorize it, we have:
\begin{align}
&R_n = \mathcal{M} e^\xi Q_n \approx \mathcal{M} (I + \xi) Q_n \quad \text{  and  } \quad \text{vec}(\xi) = \alpha^{+}\beta, \nonumber \\
&\alpha = \left(\sum_{p=1}^{P} (S_{n,p}S_{n,p}^T + \a S_{n,p}S_{n,p}^T\a^T)^T Q_n^T \right) \otimes \mathcal{M}, \label{supp_R_R} \\
&\beta = \v\left(\sum_{p=1}^{P} (Y_{n,p}S_{n,p}^T + Y_{n,p}^{\dag}S_{n,p}^T\a^T) - Q_n\sum_{p=1}^{P} (S_{n,p}S_{n,p}^T + \a S_{n,p}S_{n,p}^T\a^T) \right) \nonumber,
\end{align}
where the subscript $p$ means the $p$th keypoint, $\alpha^+$ means the pseudo inverse matrix of $\alpha$, $\otimes$ denotes Kronecker product.

\end{document}